%% file: arxiv.tex
\documentclass[11pt]{article}
\usepackage[]{acl}

\usepackage{times}
\usepackage{subfig}
\usepackage{latexsym}
\usepackage{multirow}
\usepackage{booktabs}
\usepackage{amsfonts,amsmath,amssymb}
\usepackage{tcolorbox}

\usepackage{url}
\usepackage[T1]{fontenc}
\usepackage[utf8]{inputenc}

\usepackage{microtype}

\usepackage{inconsolata}

\usepackage{graphicx}

\definecolor{llmDefault}{RGB}{0,82,163}   % Darker blue
\definecolor{llmEnhanced}{RGB}{0,122,61}   % Darker green

{%
  \begin{tcolorbox}[left=1.5mm, right=1.5mm, top=1.5mm, bottom=1.5mm]
    \raggedright
    \small
    \ifx\relax#1\relax
    \else
      \begin{center}
        {\normalsize \textbf{\color{black} #1}}
      \end{center}
    \fi
    \begin{tabular}{p{\linewidth}}
      \textcolor{black}{\fontsize{8.3}{7}\textbf{Problem:}{\fontsize{8.3}{7}\texttt{#2}}} \\
    \end{tabular}
    \vspace{1.5mm}
    \hrule
    \vspace{1.5mm}
    \begin{tabular}{p{0.45\linewidth}|p{0.45\linewidth}}
      \textcolor{llmDefault}{\fontsize{8.3}{7}\textbf{Generation (no steering):} {\fontsize{8.3}{7}\texttt{#3}}}
      &
      \textcolor{llmEnhanced}{\fontsize{8.3}{7}\textbf{Generation (steering):} {\fontsize{8.3}{7}\texttt{#4}}}
    \end{tabular}
}
{%
  \end{tcolorbox}
}

\newenvironment{promptbox}[4][] % [Title (optional)]{Prompt}{Baseline}{Intervention}
{
  \begin{tcolorbox}[left=1.5mm, right=1.5mm, top=1.5mm, bottom=1.5mm]
    \raggedright
    \small
    \ifx\relax#1\relax\else
      \begin{center}
        {\normalsize \textbf{\color{black} #1}}
      \end{center}
    \fi
    \textcolor{black}{\fontsize{8.3}{7}\textbf{Problem:}{\fontsize{8.3}{7}\texttt{#2}}} \\[2pt]
    \textcolor{llmDefault}{\fontsize{8.3}{7}\textbf{Generation (no steering):} {\fontsize{8.3}{7}\texttt{#3}}} \\[2pt]
    \textcolor{llmEnhanced}{\fontsize{8.3}{7}\textbf{Generation (steering):} {\fontsize{8.3}{7}\texttt{#4}}}
  \end{tcolorbox}
}{}

\newcommand{\score}{\texttt{ReasonScore}}

\title{I Have Covered All the Bases Here: Interpreting Reasoning Features in Large Language Models via Sparse Autoencoders}

\author{
 \textbf{Andrey Galichin\textsuperscript{1,2,3}},
 \textbf{Alexey Dontsov\textsuperscript{1,5}}, \\
 \textbf{Polina Druzhinina\textsuperscript{1,3}}, 
\textbf{Anton Razzhigaev\textsuperscript{1,3}}, \\
 \textbf{Oleg Y. Rogov\textsuperscript{1,2,3}},
 \textbf{Elena Tutubalina\textsuperscript{1,4},
\textbf{Ivan Oseledets\textsuperscript{1,3}}
 }
\\
 \textsuperscript{1}AIRI
 \textsuperscript{2}MTUCI
 \textsuperscript{3}Skoltech
 \textsuperscript{4}Sber
 \textsuperscript{5}HSE
\\
 \small{
   \textbf{Correspondence:} \href{mailto:galichin@airi.net}{galichin@airi.net}; \href{mailto:rogov@airi.net}{rogov@airi.net}
 }
}

\begin{document}
\maketitle

\begin{abstract}
Recent LLMs like \textsc{DeepSeek-R1} have demonstrated state-of-the-art performance by integrating deep thinking and complex reasoning during generation. However, the internal mechanisms behind these reasoning processes remain unexplored. We observe \textit{reasoning} LLMs consistently use vocabulary associated with human reasoning processes. We hypothesize these words correspond to specific reasoning moments within the models' internal mechanisms. To test this hypothesis, we employ Sparse Autoencoders (SAEs), a technique for sparse decomposition of neural network activations into human-interpretable features. We introduce \textit{ReasonScore}, an automatic metric to identify active SAE features during these reasoning moments. We perform manual and automatic interpretation of the features detected by our metric, and find those with activation patterns matching uncertainty, exploratory thinking, and reflection. Through steering experiments, we demonstrate that amplifying these features increases performance on reasoning-intensive benchmarks ($+2.2\%$) while producing longer reasoning traces ($+20.5\%$). Using the model diffing technique, we provide evidence that these features are present only in models with reasoning capabilities. Our work provides the first step towards a mechanistic understanding of reasoning in LLMs.\footnote{Code available at \url{https://github.com/AIRI-Institute/SAE-Reasoning}}
\end{abstract}

\section{Introduction}

\begin{figure}[t!]
\centering
\begin{promptbox}{
Convert the point $(0,3)$ in rectangular coordinates to polar coordinates.  Enter your answer in the form $(r, \theta),$ where $r > 0$ and $0 \le \theta < 2 \pi.$
}{
Okay, so I have this problem where I need to convert the rectangular coordinates (0, 3) to polar coordinates. \\ $\{... \ \textbf{1500} \ \text{tokens} \ ...\}$ $\leftarrow$ \textbf{Reasoning trace} \\ So, putting it all together, (0, 3) in rectangular coordinates is $\boxed{(3, \pi/2)}$ in polar coordinates.
}{
Okay, so I have this problem here where I need to convert the point (0, 3) from rectangular (which is the same as Cartesian) coordinates to polar coordinates. \\ $\{... \ \textbf{2000} \ \text{tokens} \ ...\}$ $\leftarrow$ \textbf{Increased reasoning trace}  \\ $\underline{\text{I think I've covered all the bases here.}}$ Calculated r, determined $\theta$, checked using different methods, and even considered the quadrant placement. I don't see any issues with the reasoning. So, I feel confident that the polar coordinates for the point (0, 3) are $\boxed{(3, \pi/2)}$.
}
\end{promptbox}
\caption{Illustration of steering (amplifying) reasoning-specific features during LLM generation. Default generation (blue) shows standard model reasoning, whereas steering (green) induces increased reasoning, self-correction, and graceful transition to the final answer—evidence that the identified features are responsible for the \textit{reasoning} concept.}
\label{fig:intro_steering}
\end{figure}

Large Language Models (LLMs) have achieved remarkable success in natural language processing \cite{brown2020language}, evolving beyond simple token prediction tasks towards explicit reasoning behaviors, such as step-by-step problem-solving \cite{wei2022chain, kojima2022large, wang2022self} and self-reflection \cite{madaan2023self, shinn2023reflexion}. Recently, specialized models which we denote as \textit{reasoning} models, such as OpenAI's o$1$ \cite{openaio1} and \textsc{DeepSeek-R1} \cite{guo2025deepseek}, have significantly improved performance on complex reasoning tasks. Trained through advanced fine-tuning and reinforcement learning \cite{shao2024deepseekmath}, these models incorporate reasoning and reflective problem-solving by generating long chains of thought before providing final answers. These advances raise a new research question: How are such reasoning capabilities internally encoded within LLMs?

A growing body of work suggests that LLMs represent meaningful concepts as linear directions in their activation spaces \cite{mikolov2013efficient, elhage2022toy, park2023linear, nanda2023emergent, jiang2024origins}. However, identifying these directions remains challenging. Sparse Autoencoders (SAEs) offer a principled approach to disentangle activations into sparse, interpretable \textit{features} \cite{cunningham2023sae, gao2024scalingSAE, templeton2024scaling, marks2024sparse}. Given a trained SAE, the interpretation of its features could be performed by activation analysis \cite{bricken2023towards}, targeted interventions \cite{templeton2024scaling}, or automated methods \cite{paulo2024automatically, kuznetsov2025feature}. While SAEs have proven effective in discovering features for various concepts \cite{shu2025survey}, their ability to isolate reasoning-specific features remains unexplored.

In this work, we investigate whether reasoning processes in \textit{reasoning} LLMs can be identified and decomposed into interpretable directions within their activation spaces. We analyze the outputs produced by these models', and find a consistent pattern in which they employ words associated with human reasoning processes: uncertainty (e.g. ``perhaps''), reflection (e.g. ``however''), and exploration (e.g. ``alternatively'') \cite{structure_of_discussions, reasoning_boyd, conginive_discource_explore}. We hypothesize that these linguistic patterns correspond to the moments of reasoning within the models' internal mechanisms. To test this, we construct a vocabulary of reasoning words. We then use SAEs to decompose LLM activations into interpretable features and propose \score{}, a metric that quantifies the degree to which a given SAE feature is active on the reasoning vocabulary.

We evaluate the features found by \score{} using manual \cite{bricken2023towards} and automatic interpretation \cite{kuznetsov2025feature} techniques, and find the set of $46$ features that demonstrate interpretable activation patterns corresponding to uncertainty, exploratory thinking, and reflection. We perform steering experiments and show that amplifying these reasoning features leads to improved performance on reasoning-intensive benchmarks ($+13.4\%$ on AIME-2024, $+2.2\%$ on MATH-500, and $+4\%$ on GPQA Diamond) while producing longer reasoning traces ($+18.5\%$ on AIME-2024, $+20.5\%$ on MATH-500, and $+13.9\%$ on GPQA Diamond). Through model diffing \cite{bricken2024stage}, we demonstrate that these reasoning features emerge only in \textit{reasoning} LLMs and are absent in base models. Our results provide mechanistic evidence that specific, interpretable components in LLMs representations are causally linked to reasoning behavior.

The contributions of this paper are the following:
\begin{itemize}
    \item We introduce \score{}, an automatic metric to identify the SAE features responsible for reasoning and confirm its effectiveness using interpretability techniques.
    \item We provide causal evidence from steering experiments, demonstrating that amplifying identified features induces reasoning behavior.
    \item We analyze the emergence of reasoning features in LLMs through model diffing technique, and confirm their existence only after the \textit{reasoning} fine-tuning stage.
\end{itemize}

%\section{Related Work}

\section{Interpretability with SAEs}

Sparse Autoencoders (SAEs) aim to learn a sparse decomposition of model activations to identify disentangled features that correspond to meaningful concepts \cite{bricken2023towards}. Here, a \textit{feature} refers to an individual component of the learned representation that captures specific, human-interpretable characteristics of the input data.

The core idea behind SAEs is to reconstruct model activations $x \in \mathbb{R}^{n}$ as a sparse linear combination of learned feature directions, where the feature \textit{dictionary} dimensionality $m \gg n$. Formally, we extract LLM activations from some intermediate state in the model and train a two-layer autoencoder:
\begin{equation}
\label{eq:sae}
\begin{aligned}
f(x) &= \sigma(W_{\text{enc}}x + b_{\text{enc}}), \\
\hat{x}(f) &= W_{\text{dec}}f + b_{\text{dec}}.
\end{aligned}
\end{equation}
Here, $f(x) \in \mathbb{R}^{m}$ is a sparse vector of feature magnitudes and $\hat{x}(f) \in \mathbb{R}^{n}$ is a reconstruction of the original activation $x$. The columns of $W_{\text{dec}}$, which we denote by $W_{\text{dec}, i}$, $i = 1, \ ..., \ m$, represent the dictionary of directions, or \textit{features}, into which the SAE decomposes $x$. The activation function $\sigma$ enforces non-negativity in $f(x)$.

The training objective used to train Sparse Autoencoders minimizes a reconstruction loss $\mathcal{L}_{\text{recon}}$ and an additional sparsity-promoting loss $\mathcal{L}_{\text{sparsity}}$. This objective forces SAE to learn a small set of interpretable features that capture the distinct properties of the activations.

In our work, we use vanilla SAE \cite{bricken2023towards} with \texttt{ReLU} activation function. Following \cite{conerly2024update}, we use a squared error reconstruction loss and a modified $\text{L}1$ penalty as a sparsity loss:
\begin{equation}
\mathcal{L} = \underbrace{\left\|x - \hat{x}\right\|_2^2}_{\mathcal{L}_\text{recon}} + \lambda \underbrace{\sum\nolimits_{i=1}^m f_i\left\|W_{\text{dec},i}\right\|_2}_{\mathcal{L}_\text{sparsity}},
\end{equation}
where $\lambda$ is the sparsity penalty coefficient.

\section{Method}

We identify reasoning-specific features through a two-step approach. First, we examine the language space of reasoning words used by \textit{reasoning} LLMs, and construct the respective vocabulary $\mathcal{R}$ (Sec.~\ref{sec:reasoning_space}). Secondly, we introduce \score{} to find the sparse autoencoder features responsible for reasoning capabilities (Sec. \ref{sec:reasoning_relevance}).
\label{sec:space}

\subsection{Reasoning Vocabulary}
\label{sec:reasoning_space}

Reasoning words are linguistic features associated with exploratory talk as humans talk-to-learn, explore ideas, and probe each other's thinking \cite{reasoning_boyd}.

\begin{figure}[t!]
    \centering
    \includegraphics[width=0.85\linewidth]{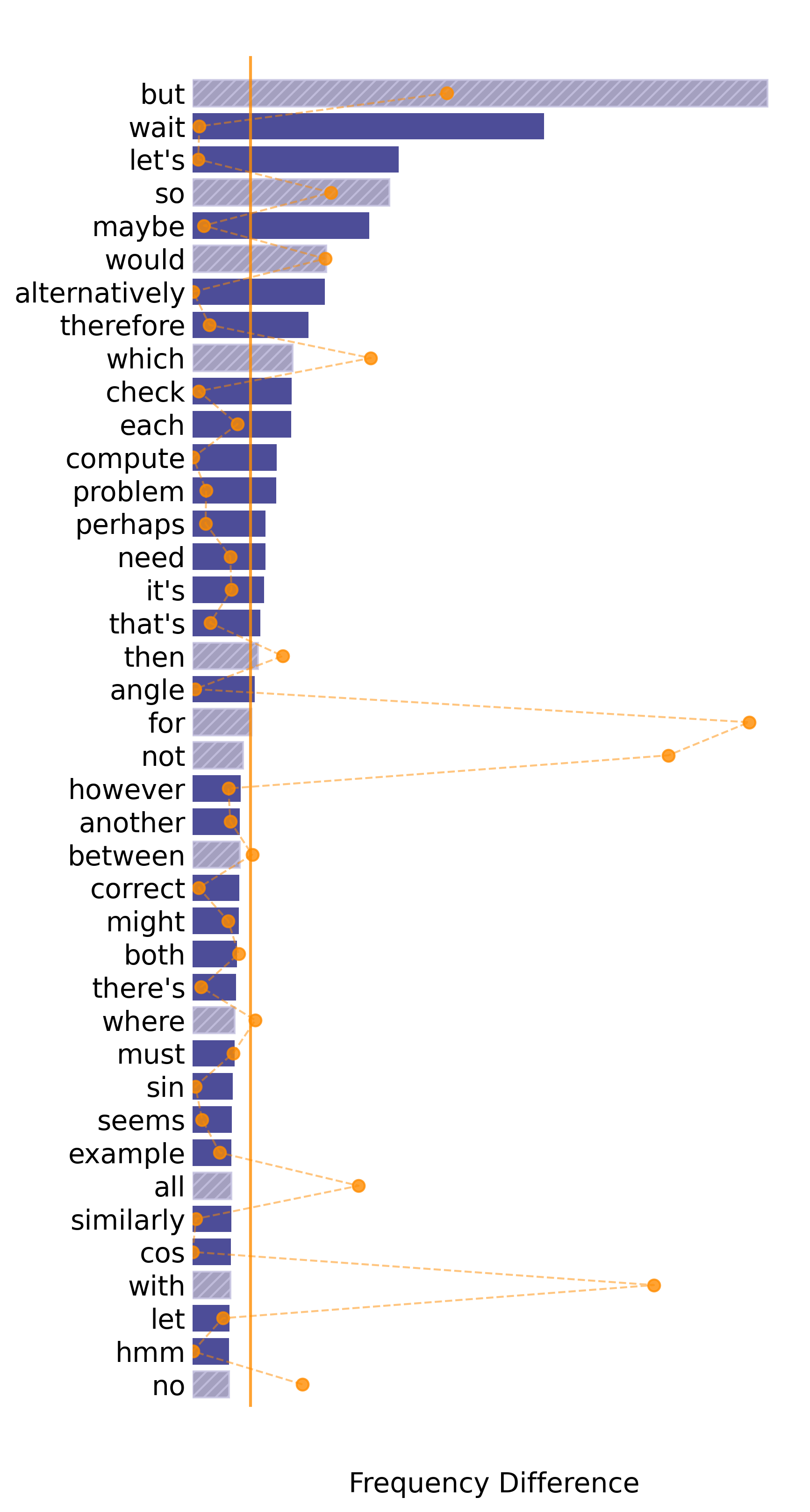}
    \caption{The distribution of top 40 words with the greatest change in frequency between reasoning traces of \textsc{DeepSeek-R1} and ground-truth solutions of math problems. Orange dots show the frequency from Google Books Ngram Corpus. We remove the words with absolute frequency above the pre-defined threshold (orange line), and keep those with the high relative frequency indicating reasoning.}
    \label{fig:words_distrib}
\end{figure}

In the original \textsc{DeepSeek-R1} paper \cite{guo2025deepseek}, the authors demonstrated that the model spontaneously exhibits sophisticated human-like behaviors, such as reflection, where it revisits and reevaluates its previous steps, and exploration of alternative problem-solving approaches. In particular, the model explicitly employs words that mirror the introspective language humans use when thinking (such as ``maybe'', ``but'', ``wait''). We hypothesize that these moments correspond directly to the internal reasoning process of the models, which is consistent with studies on human thinking \cite{structure_of_discussions, reasoning_boyd}.

To extract the models' reasoning vocabulary, we use an approach similar to that of \cite{frequency_profiling}. We construct two corpora from the \textsc{OpenThoughts-114k} \cite{openthoughts} dataset: ground-truth samples containing formal and step-by-step solutions to the problems, and the solutions obtained using \textsc{DeepSeek-R1} for the same problems. For each word, we calculate its frequency in the tasks solutions $p_{\text{solution}}$ and in the thinking solutions $p_{\text{think}}$, then sort all words by the frequency difference $p_{\text{think}} - p_{\text{solution}}$. Next, we select the top-$k$ words by frequency difference, where $k$ is determined by the point where the frequency distribution plateaus, and filter out words with high presence in the Google Books Ngram Corpus \cite{michel2011quantitative} (Fig.~\ref{fig:words_distrib}). To determine the final vocabulary from this candidate set, we choose words that best capture reasoning behavior. This includes words that match those considered in the linguistic literature \cite{structure_of_discussions, reasoning_boyd} and those that we identify through manual analysis of model traces exhibiting reasoning patterns.

Following this pipeline, we select $10$ words indicating reasoning as models' \textit{reasoning} vocabulary and denote it by $\mathcal{R}$. The exact list of words can be found in Appx.~\ref{sec:appendix_reasoning_vocabulary}. Ablation experiments confirm that these words play a functional role in reasoning capabilities (see Sec.~\ref{sec:steering} for setup, Appx.~\ref{sec:appendix_functional_importance} for results).

\subsection{ReasonScore}
\label{sec:reasoning_relevance}

\begin{figure*}[t!]
    \centering
    \subfloat[Top-activating examples from the manually verified set of features.]
    {
    \label{fig:main_interface}
    \includegraphics[width=0.48\linewidth]{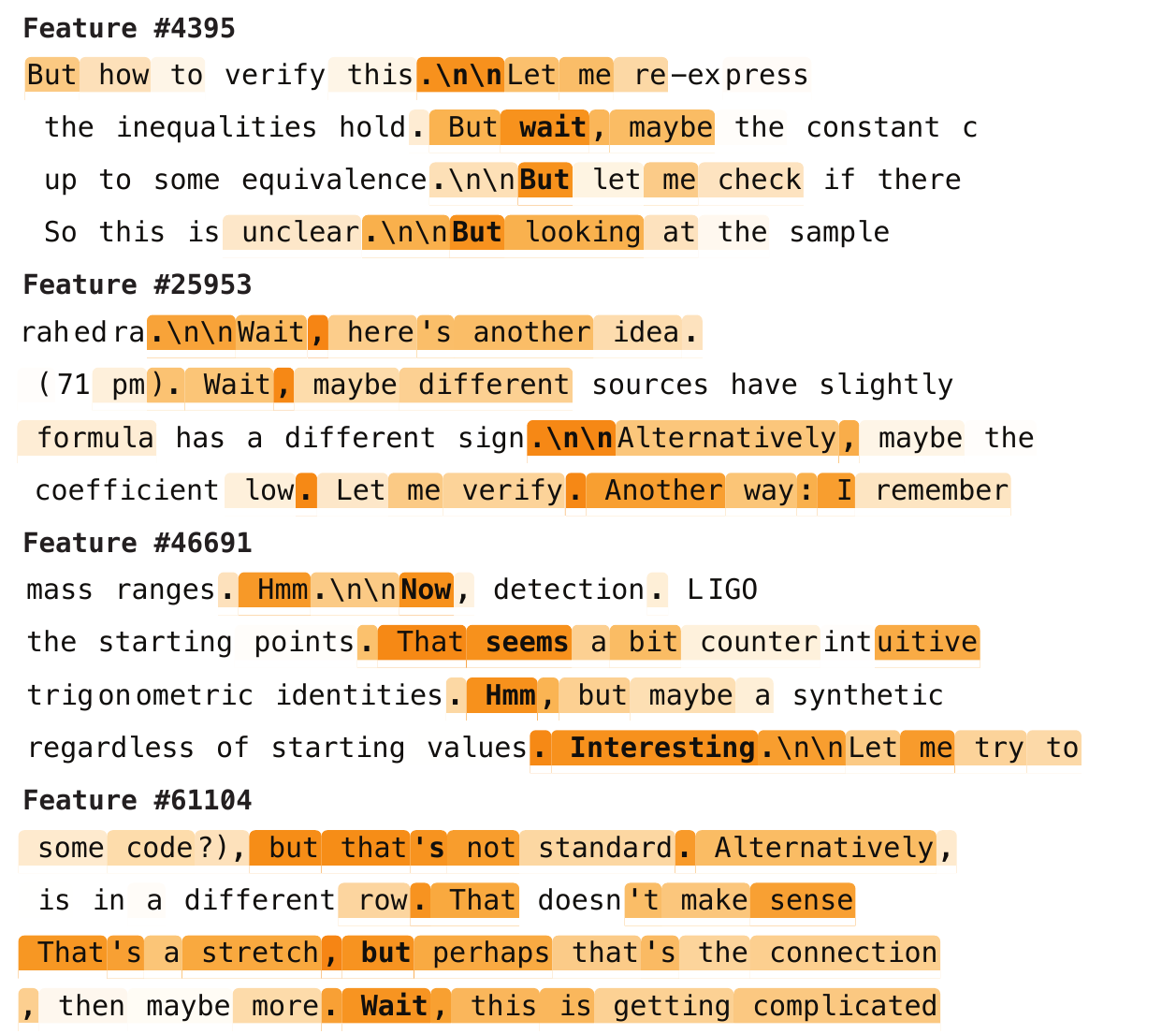}
    }\vspace{1em} 
    \subfloat[Distribution of manually verified set of features on function groups generated by \textsc{GPT-4o}.]
    {
    \label{fig:main_autointerp}
    \includegraphics[width=0.44\linewidth]{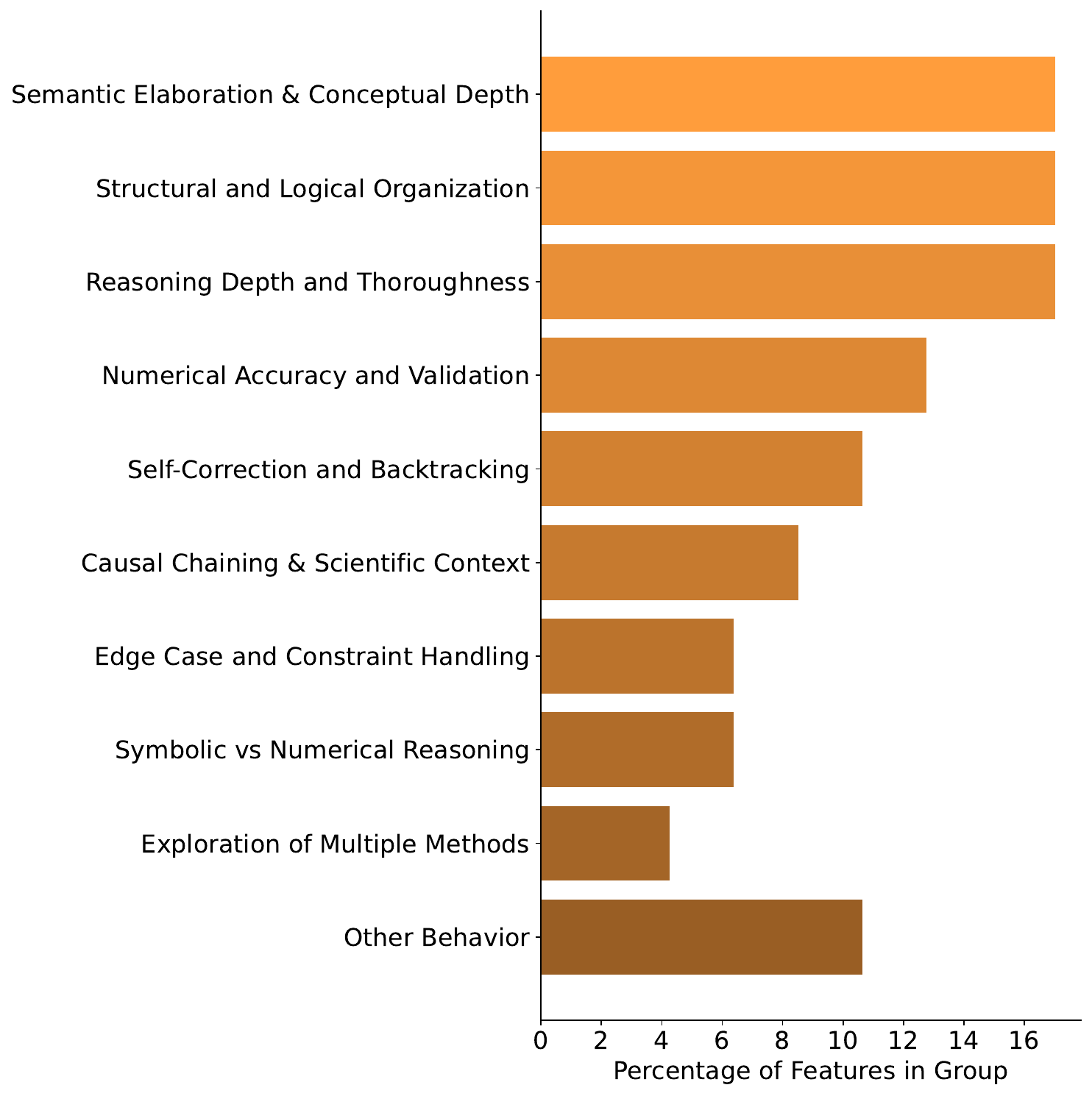}
    }
    \caption{Interpretability results for manually verified set of features in our SAE: (a) Examples of feature interfaces used in manual interpretation experiments, (b) Distribution of reasoning features on function groups obtained by automatic interpretation pipeline by using \textsc{GPT-4o} as a judge.}
    
    % \caption{Empirical analysis results for the feature $17456$ in our SAE: (a) Distribution of the bottom and top logits influenced by one of the ``reasoning'' features. (b) Contexts where the ``reasoning'' feature activates the most. Increased color intensity represent higher activation values, with token highlighted in bold having the highest activation.}
    \label{fig:interpretation}
\end{figure*}

To find SAE features that capture reasoning-related behavior, we follow our hypothesis and introduce \score{}, which measures the contribution of $i$-th feature to reasoning. Using a dataset of model's activations (see details in Sec. \ref{sec:setup_data}) $\mathcal{D} = \mathcal{D_R} \cup \mathcal{D_{\neg R}}$, where $\mathcal{D_R}$ contains token activations corresponding to words in $\mathcal{R}$ and $\mathcal{D_{\neg R}}$ contains all other activations, we first define a score:
\begin{equation}
\label{eq:score}
s_i = \frac{\mu(i, \mathcal{D_R})}{\sum_j \mu(j, \mathcal{D_R})} - \frac{\mu(i, \mathcal{D_{\neg R}})}{\sum_j \mu(j, \mathcal{D_{\neg R}})},
\end{equation}
where $\mu(i, \mathcal{D}) = \frac{1}{|\mathcal{D}|}\sum_{x \in \mathcal{D}}f_i(x)$ is the average activation value of the $i$-th feature on dataset $\mathcal{D}$. This score is similar to the one in \cite{cywinski2025saeuron} and identifies features that concentrate the most of their activation mass on reasoning words. 

However, analysis of feature activations only on individual words may miss important contextual information. The words in $\mathcal{R}$ are critical indicators of the reasoning process and also serve as transition points, signaling shifts in the thought process, uncertainty, or reflection. Therefore, a feature involved in reasoning should activate not only on the reasoning words, but also as the model approaches and continues through these transitions. To capture it, we define $\mathcal{D}_\mathcal{R}^{\text{W}}$ as the dataset that contains activations within a fixed-width context window around tokens corresponding to words in $\mathcal{R}$, and $\mathcal{D}_\mathcal{\neg R}^{\text{W}}$ contains all other activations. We modify Eq.~\ref{eq:score} to use the new version of the datasets.

To penalize features that activate only on a small fraction of $\mathcal{R}$, we further introduce an \textit{entropy penalty}. For $i$-th feature, we first calculate $\mu(i, \mathcal{D}_{r_j}^{\text{W}})$ for each word $r_j \in \mathcal{R}$, normalize these values into a probability distribution $p_i(r_j) = \frac{\mu(i, \mathcal{D}_{r_j}^{\text{W}})}{\sum_{k \in \mathcal{R}}\mu(i, \mathcal{D}_{r_k}^{\text{W}})}$, and compute the entropy:
\begin{equation}
\label{eq:entropy}
\mathrm{H}_i = -\frac{1}{\log|\mathcal{R}|} \cdot \sum\limits_{j}^{|\mathcal{R}|}p_i(r_j)\log p_i(r_j).
\end{equation}
Here, $\log|\mathcal{R}|$ normalizes the entropy to $[0,1]$, with $\mathrm{H}_i = 1$ indicating perfect uniformity over $\mathcal{R}$. By adding the entropy penalty in Eq.~\eqref{eq:score}, we define the \score{} for the $i$-th SAE feature as:
\begin{equation}
\label{eq:score_w_entropy}
\begin{split}
\text{\score}_i ={} & \frac{\mu(i, \mathcal{D_R^\text{W}})}{\sum_j \mu(j, \mathcal{D_R^\text{W}})} \cdot \mathrm{H}_i^\alpha \\
                   & - \frac{\mu(i, \mathcal{D_{\neg R}^\text{W}})}{\sum_j \mu(j, \mathcal{D_{\neg R}^\text{W}})}.
\end{split}
\end{equation}
where $\alpha$ controls the trade-off between specificity ($\alpha \rightarrow 0$) and generalization ($\alpha > 1$).

We identify the set of reasoning features in a SAE based on their \score{} and define the corresponding set of feature indices as:
\begin{equation}
\label{eq:reason_indices}
\mathcal{F}_\mathcal{R} = \{i \mid i \in [1, m], \ \text{\score}_i > \tau  \},
% \mathcal{F}_\mathcal{R} = \{i_j \mid \forall j \in [1, m]: \text{RS}_{i_j} \geq \text{RS}_{i_{j+1}}\},
\end{equation}
where $\tau$ is the $q$-th quantile of the \score{} distribution across all features.

\section{Evaluation}

In this section, we analyze how effectively our discovered features model reflection, uncertainty, and exploration within the \textit{reasoning} model. We discuss our experimental setup (Sec.~\ref{sec:experimental_setup}), perform manual and automatic interpretation of the features we find (Sec.~\ref{sec:interpretability}), and conduct steering experiments with these features on various benchmarks (Sec.~\ref{sec:steering}). Finally, we apply the model diffing technique to demonstrate that these features exist only in models with reasoning capabilities (Sec.~\ref{sec:diffing}).

\subsection{Experimental Setup}
\label{sec:experimental_setup}

% Here, we describe how we train the Sparse Autoencoder to obtain sparse decomposition of the LLM activations into interpretable features.
\paragraph{Model.}

We apply SAE to the output activations from the $19$-th layer of the \textsc{DeepSeek-R1-Llama-8B} model. This model was selected for its \textit{reasoning} capabilities and open-source availability. The $19$-th layer ($\approx 60\%$ model depth) was chosen because at this point LLMs predominately store the most of their knowledge \cite{chen2023beyond, jin2024exploring}. We provide results for other layers of \textsc{DeepSeek-R1-Llama-8B} and another model family in Appx.~\ref{sec:appendix_generalization}.

\paragraph{Data.}
\label{sec:setup_data}

We train SAE on the activations of the model generated using text data from the \textsc{LMSys-Chat-1M} \cite{zheng2023lmsyschat1m} and \textsc{OpenThoughts-114k} \cite{openthoughts} datasets. The first provides a broad and diverse spectrum of real-world text data, which we denote as \textit{base data}, while the latter provides high-quality reasoning traces generated by \textsc{DeepSeek-R1} for math, science, code, and puzzle samples, which we denote as \textit{reasoning data}. The SAE is trained on $1$B tokens, evenly split between the two datasets, with a context window of $1{,}024$ tokens.

\paragraph{Training.}

We set the SAE dictionary dimensionality to $m = 65{,}536$, which is $16$ times larger than the model activation size $n = 4{,}096$ following established practices \cite{lieberum2024gemma}, and adopt the same training settings as in the Anthropic April update \cite{conerly2024update}. We train with the Adam optimizer \cite{kingma2014adam} with $(\beta_1, \ \beta_2) = (0.9, 0.999)$, batch size of $4{,}096$, and a learning rate $\eta = 5 \times 10^{-5}$. The learning rate is decayed linearly to zero over the last $20\%$ of training. The gradient norm is clipped to $1$. We use a linear warmup for the sparsity coefficient from $\lambda = 0$ to $\lambda = 5$ over the first $5\%$ training steps.

\paragraph{Evaluation.}

We use the mean $\text{L}0$-norm of latent activations, $\mathbb{E}_x\|f(x)\|_0$, as a measure of sparsity. To measure reconstruction quality, we use fraction of variance of the input explained by the reconstruction. Both metrics were computed on $2{,}048$ sequences of length $1{,}024$.

At a $\text{L}0$ of $86$ the reconstruction of our SAE explains $68.5\%$ of the variance in model activations. This shows that our SAE achieves reliable reconstruction performance at a low sparsity level, allowing a decomposition of raw activations into interpretable features.

\paragraph{\score.}

We calculate \score{} (Eq.~\ref{eq:score_w_entropy}) on $10$M tokens from the \textsc{OpenThoughts-114k} dataset. To collect $\mathcal{D}_\mathcal{R}^{\text{W}}$, we use an asymmetric window with $2$ preceding and $3$ subsequent tokens, following established practices in Keyphrase extraction~\cite{mihalcea2004textrank, breidt1996extraction, zhang2020empirical} We set $\alpha = 0.7$ for the \textit{entropy penalty} as a reasonable default. Based on the empirical analysis of \score{} distribution (see Appx.~\ref{sec:appendix_score_distrinution}), we set $q = 0.997$ in Eq.~\eqref{eq:reason_indices}, resulting in $|\mathcal{F}_\mathcal{R}| = 200$ features.

\begin{table}[t!]
    \centering
    \resizebox{\linewidth}{!}{
    \begin{tabular}{@{}l *{6}{c} @{}}
    \toprule
    \multirow{3}{*}{\centering\textbf{Feature \#}} & \multicolumn{2}{c}{\multirow{2}{*}{\textbf{AIME 2024}}} & \multicolumn{2}{c}{\multirow{2}{*}{\textbf{MATH-500}}} & \multicolumn{2}{c}{\textbf{GPQA}} \\
    &  &  &  &  & \multicolumn{2}{c}{\textbf{Diamond}} \\
    \cmidrule(lr){2-3} \cmidrule(lr){4-5} \cmidrule{6-7}
    & maj@4 & tokens (K) & maj@4 & tokens (K) & maj@4 & tokens (K) \\
    \midrule
    \textbf{No steering} & 53.3 & 12.4 & 93.2 & 3.9 & 50.0 & 7.9 \\
    \midrule
    \textbf{3942} & 56.7 & \underline{11.1} & 93.0 & \underline{3.4} & 46.5 & \underline{6.7} \\
    \textbf{4395} & 56.7 & \textbf{14.7} & \textbf{95.4} & 4.1 & 52.0 & 8.5 \\
    \textbf{16441} & 60.0 & 14.0 & 95.0 & 4.1 & 54.0 & 8.3 \\
    \textbf{16778} & 56.7 & 14.1 & 94.0 & \textbf{4.7} & 51.0 & \textbf{9.0} \\
    \textbf{25953} & 60.0 & 12.8 & 94.2 & 4.2 & 53.0 & 8.1 \\
    \textbf{46691} & 56.7 & 14.0 & 94.2 & 4.2 & \textbf{54.0} & 8.0 \\
    \textbf{61104} & \textbf{66.7} & 12.0 & 95.0 & 3.6 & 53.0 & 7.5 \\
    \bottomrule
    \end{tabular}
    }
    \caption{Performance and average number of output tokens for different steering experiments on reasoning-related benchmarks.}
    \label{tab:benchmarking}
\end{table}

\subsection{Interpretability of Reasoning Features}
\label{sec:interpretability}

\paragraph{Manual Interpretation.} 

To evaluate the features we found with \score{}, we manually interpret each feature from $\mathcal{F}_{\mathcal{R}}$ ($200$ in total). For each feature, we find the examples in a subset of the \textsc{OpenThoughts-114k} corpus that caused the feature to activate, and construct the interface proposed in \cite{bricken2023towards}. This mainly includes examples of when the feature activates, its effect on the logits when it does, and other statistics. We determine whether a feature qualifies as a good reasoning candidate if: (1) when it is active, the relevant concept is reliably present in the context, (2) it triggers in various examples of reasoning tasks, and (3) its activation impacts interpretable logits that correspond to reasoning processes.

Through our analysis, we identify three behavioral modes that characterize models' reasoning process:
\begin{itemize}
    \item \textbf{Uncertainty}: Moments where the model exhibits hesitation, doubts, and provisional thinking
    \item \textbf{Exploration}: Moments where the model considers multiple possibilities, connects ideas, examines different perspectives
    \item \textbf{Reflection}: Moments where the model revisits and reevaluates its previous steps
\end{itemize}
Our manual analysis reveals a set of $46$ features that exhibit these patterns, which we believe are responsible for the reasoning mechanisms of the model. We denote this set by $\mathcal{F}_\mathcal{R}^{\text{manual}} \subset \mathcal{F}_\mathcal{R}$. In Fig.~\ref{fig:main_interface}, we provide examples of feature interfaces used for interpretation. The results demonstrate features that consistently activate in contexts representing model's uncertainty ($\# 61104$), exploration ($\# 25953$), and reflection ($\# 4395,\# 46691$). Additional examples of interfaces can be found in Appx.~\ref{sec:appendix_feature_interfaces}.

\paragraph{Automatic Interpretation.}

To complement our manual analysis, we annotate these features with an automatic interpretation pipeline \cite{kuznetsov2025feature}. This approach employs feature steering, a technique that modulates feature activations to analyze their functional influence. For each $i$-th feature, we estimate its maximum activation $f_i^{\max}$ using a subset of the \textsc{OpenThoughts-114k} corpus. During response generation, we modify model activations as follows:
\begin{equation}
\label{eq:steering}
x' = x + \gamma f_i^{\max} W_{\text{dec},i},
\end{equation}
where $\gamma$ controls the steering strength.

To evaluate the impact of $i$-th feature on reasoning capabilities, we generate multiple outputs by varying $\gamma \in [-4, 4]$, pass them to \textsc{GPT-4o}, and ask it to generate an explanation or function that best describes the semantic influence caused by steering a feature. The result, shown in Fig.~\ref{fig:main_autointerp}, reveals that the features we found group into distinct reasoning-related patterns. Only a small fraction of features from $\mathcal{F}_\mathcal{R}^{\text{manual}}$ ($5$) was assigned to a mixed class ``Other Behavior'' containing mixed explanation. We provide a more comprehensive description of auto-interpretability pipeline results in Appx.~\ref{sec:appendix_autointerp}.

\begin{tcolorbox}
\textbf{Takeaway 1}: Manual interpretation experiments confirm that \score{} identifies features that describe model's reasoning capabilities, revealing $46$ features that represent uncertainty, exploration, and reflection. Automatic interpretation demonstrates that these features are causally linked to reasoning behavior.
\end{tcolorbox}

\subsection{Steering Reasoning Features}
\label{sec:steering}

\begin{figure}[t]
\centering
    \includegraphics[width=1.0\linewidth]{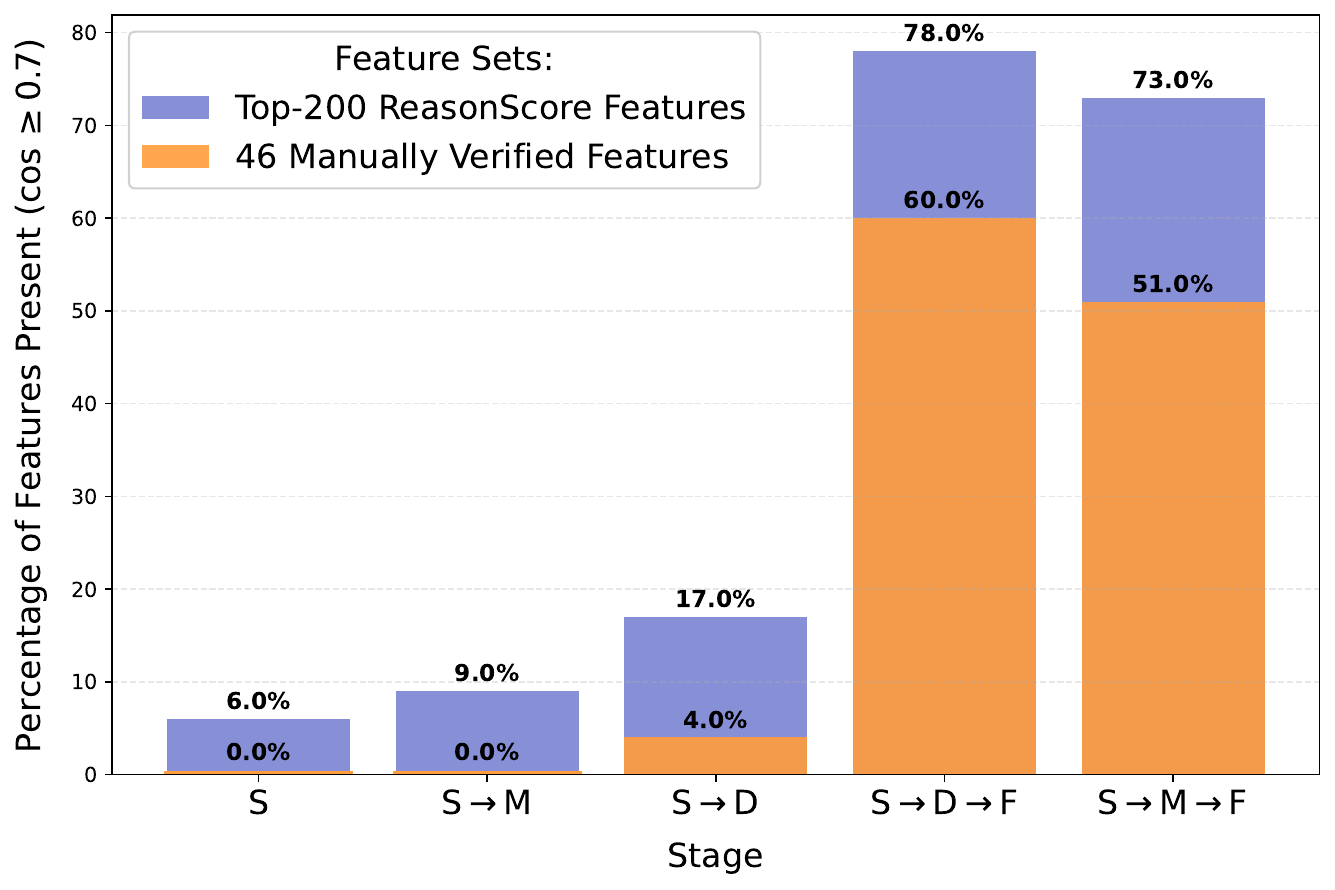}
    \caption{Percentage of \score{} features present at each stage of the diffing pipeline. The blue bars represent the features from $\mathcal{F}_\mathcal{R}$, the orange bars represent the $\mathcal{F}_\mathcal{R}^{\text{manual}}$ features. Features are considered present if their cosine similarity with any feature in corresponding stage's SAE is $\geq 0.7$. Stages: (\textbf{S}) Base model + base data; (\textbf{S$\rightarrow$D}) Base model + reasoning data; \textbf{S$\rightarrow$M} Reasoning model + base data; (\textbf{S$\rightarrow$D/M$\rightarrow$F}) Reasoning model + reasoning data.}
    \label{fig:feature_emergence}
\end{figure}

To demonstrate whether our interpretations of features describe their influence on model behavior, we further experiment with feature steering.

Our goal is to verify if steering reasoning features improve the LLM's performance on reasoning-related benchmarks. Following the setup in \textsc{DeepSeek-R1}, we evaluate performance on AIME 2024 \cite{aime}, MATH-500 \cite{hendrycks2021measuringmathematicalproblemsolving}, and GPQA Diamond \cite{rein2023gpqagraduatelevelgoogleproofqa}. To obtain steering results for $i$-th feature, we modify the activations during response generation according to Eq.~\eqref{eq:steering}. To determine the optimal steering strength that can influence model outputs without significantly damaging capabilities, we ran evaluations with a small subset of $10$ reasoning features on MATH-500. We varied the steering strength $\gamma$ from $1$ to $8$. Based on these experiments, we determined the optimal range $\gamma \in [1, 3]$, which aligns with the findings in \cite{durmusevaluating}. For all subsequent experiments, we set the steering strength $\gamma = 2$.

We perform a preliminary analysis to identify the most promising features for reasoning enhancement from our set of manually chosen features $\mathcal{F}_\mathcal{R}^{\text{manual}}$. For each feature, we measure the accuracy (or $\text{pass}@1$ \cite{chen2021evaluating}) on MATH-500 and evaluate the results. Of the $46$ features, $9$ improve performance by $\geq 0.5\%$, $29$ show no or minimal performance degradation ($\leq 2.0$), and the remaining $8$ decrease performance by at most $4\%$. Interestingly, we identify feature $\# 3942$, which produces substantially shorter responses while maintaining negligible performance degradation. For further analysis, we select the $9$ top-performing features and feature $\# 3942$.

We evaluate these $10$ features across all reasoning benchmarks. We report $\text{maj}@4$ (\cite{wang2022self, muennighoff2025s1}) and the average number of tokens generated during the model's thinking process. The results, shown in Tab.~\ref{tab:benchmarking}, demonstrate that steering $7$ out of $10$ features produces consistent improvements in both performance and reasoning depth. Feature $\# 4395$ yields the most significant performance gain on MATH-500 ($+2.2\%$). Feature $\# 16778$ produces the longest reasoning traces on average ($+13.7\%$ on AIME-2024, $+20.5\%$ on MATH-500, and $+13.9\%$ on GPQA Diamond) and consistently outperforms the ``no steering'' baseline. Feature $\# 3942$ produces shortest reasoning traces on average ($-7.7\%$) with minor performance degradation. We provide examples of generated solutions without and with feature steering in Appx.~\ref{sec:appendix_steering}.

\begin{tcolorbox}
\textbf{Takeaway 2}: 
We find that amplifying certain reasoning features prolongs the internal thought process and correlates with increased performance on reasoning-related tasks.
\end{tcolorbox}

\subsection{Stage-wise Emergence of Reasoning Features}
\label{sec:diffing}

Our interpretation experiments (Sec.~\ref{sec:interpretability}) revealed that features identified by \score{} exhibit activation patterns consistent with reasoning processes. The steering experiments (Sec.~\ref{sec:steering}) provided causal evidence by demonstrating that amplification of these features improves performance on reasoning-intensive benchmarks. Given these findings, we now aim to answer the next important question: do these reasoning features naturally emerge during standard pre-training procedure, or are they specifically induced by the \textit{reasoning} fine-tuning process? 

To answer this question, we use the stage-wise fine-tuning technique proposed in \cite{bricken2024stage}. This approach aims to isolate how features evolve across different model and dataset combinations. In our experiments, we examine how the features change between two model states: before (\textit{base model}) and after (\textit{reasoning model}) \textit{reasoning} fine-tuning stage. We accomplish this by training a SAE on the base model before it has been fine-tuned, and then fine-tuning it on either the reasoning model or the fine-tuning data. Formally, we define four distinct stages:
\begin{description}
    \item[Stage S:] base model + base data (starting point)
    \item[Stage D:] base model + reasoning data (isolating dataset effects)
    \item[Stage M:] reasoning model + base data (isolating model effects)
    \item[Stage F:] reasoning model + reasoning data (full fine-tuning)
\end{description}
We analyze these changes through two fine-tuning trajectories, each involving two sequential fine-tuning stages: (1) \textbf{S}$\rightarrow$\textbf{D}$\rightarrow$\textbf{F} takes initial SAE (Stage \textbf{S}), fine-tunes it on reasoning data (\textbf{S}$\rightarrow$\textbf{D}), and finally fine-tunes on both reasoning model and reasoning data (\textbf{D}$\rightarrow$\textbf{F}); (2) \textbf{S}$\rightarrow$\textbf{M}$\rightarrow$\textbf{F} takes initial SAE (Stage \textbf{S}), fine-tunes it on reasoning model (\textbf{S}$\rightarrow$\textbf{M}), and finally fine-tunes on both reasoning model and reasoning data (\textbf{M}$\rightarrow$\textbf{F}). If reasoning features are present only in \textit{reasoning} models, we should observe the emergence of these features in response to \textbf{both} reasoning model and reasoning data (Stage \textbf{F}). This corresponds to the final steps of the fine-tuning trajectories: (\textbf{S}$\rightarrow$\textbf{D}/\textbf{M}$\rightarrow$\textbf{F}).

We use \textsc{Llama-3.1-8B} \cite{grattafiori2024llama} as base model and \textsc{SlimPajama} \cite{cerebras2023slimpajama} as base data. We select \textsc{SlimPajama} over \textsc{LMSys-Chat-1M} as our \textit{base data} because it better matches the pre-training distribution of \textsc{Llama-3.1-8B}, which has not undergone instruction-tuning. For each stage, we use the same setup as in Sec.~\ref{sec:experimental_setup}, with each fine-tuning stage taking $30\%$ of the total number of tokens required for training from scratch. For each $i$-th feature from $\mathcal{F}_\mathcal{R}$, we check its existence at each stage by measuring cosine similarity ($\cos$) between feature vectors. We follow \cite{bricken2024stage} and consider a feature present if $\cos \geq 0.7$ with any feature in a SAE of the corresponding stage.

Fig.~\ref{fig:feature_emergence} shows the percentage of reasoning features present at each fine-tuning stage. We find that the reasoning features are almost absent in the base model and after switching to the reasoning model ($0\%$ of manually verified features $\mathcal{F}_\mathcal{R}^{\text{manual}}$). When introducing the reasoning data to the base model (\textbf{S$\rightarrow$D}), only $4\%$ of the verified reasoning features emerge, indicating that exposure to the reasoning content alone is insufficient to develop these features. Finally, when we incorporate both the reasoning data and the reasoning model, we observe that $60\%$ of the verified reasoning features appear in the (\textbf{S}$\rightarrow$\textbf{D}$\rightarrow$\textbf{F}) stage and $51\%$ in the (\textbf{S}$\rightarrow$\textbf{M}$\rightarrow$\textbf{F}) stage. The noticeable increase in the presence of features only when both reasoning data and model are combined provides compelling evidence that \score{} identifies features associated with the model's reasoning processes rather than general capabilities.

\begin{tcolorbox}
\textbf{Takeaway 3}: We show that most of the features found by \score{} emerge only after the \textit{reasoning} fine-tuning stage. Exposure to the reasoning data or reasoning model alone is insufficient to develop these features.
\end{tcolorbox}

\section{Related Work}

\subsection{Mechanistic Interpretability}

 Various methods exist to shed light on the inner workings of LLMs, including attention analysis, which examines the model's focus on input tokens \cite{vaswani2017attention}, and gradient-based methods that identify influential input features \cite{SimonyanVZ13}. Probing techniques offer insights into the information captured within different layers of an LLM \cite{alain2016understanding}. Mechanistic interpretability aims to reverse-engineer the computations of LLMs, employing techniques like activation patching \cite{meng2022locating} and feature steering \cite{cao2024personalized, soo2025steering} to understand and control model behavior. The logit lens provides a way to observe the model's token predictions at different processing stages \cite{nostalgebraist2020logitlens}. 

\subsection{Sparse Autoencoders}

SAEs have emerged as a key tool for understanding the internal representations of LLMs in interpretability research \cite{gao2024scaling, huben2024sparse}. By learning a sparse decomposition of model activations, SAEs identify disentangled features that correspond to meaningful concepts \cite{marks2024sparse}.

SAE features are significantly more monosemantic than individual neurons, making them effective for mechanistic interpretability \cite{leask2025canonical}. A key challenge in using SAEs for interpretability is ensuring that the extracted features are monosemantic and robust. Yan et al. \cite{yan2024monosemanticity} propose using feature decorrelation losses to enforce better separation between learned latents, preventing redundancy. Recent advances in cross-layer SAEs \cite{shi2025routeSAE} enable the analysis of more abstract, high-level reasoning patterns across multiple layers.

SAEs have proven valuable for studying model development across training stages. Crosscoders \cite{lindsey2024sparse} enable direct feature mapping to model states, while stage-wise model diffing \cite{bricken2024stage} compares SAEs trained on different checkpoints. We adopt the diffing approach for its computational efficiency and intuitive implementation. While previous work has applied diffing to sleeper agents, our research extends this approach to investigate reasoning behavior.

\subsection{Reasoning LLMs}

Recent LLM innovations have focused on models with explicit reasoning abilities, including OpenAI's o$1$ \cite{openaio1}, \textsc{DeepSeek-R1} \cite{guo2025deepseek}, and \textsc{QwQ-32B-preview} \cite{team2024qwq}. These methods employ rule-based reinforcement learning using correctness scores (final answer accuracy) and format scores (output structure compliance), leading to advanced reasoning behaviors like self-correction and reflection, denoted as an ``aha moment'' in the \textsc{DeepSeek-AI} report \cite{guo2025deepseek}.

Despite the success of rule-based reinforcement learning in enabling reasoning capabilities, how these models encode their internal reasoning remains unclear. We address this problem using SAEs to find interpretable features responsible for underlying reasoning mechanisms, which, to the best of our knowledge, has not been done yet.
\section{Conclusion}

In this work, we present a novel methodology for uncovering reasoning mechanisms in LLMs through SAEs. We introduce \score{}, a metric that identifies reasoning-related SAE features based on their activation patterns using a curated introspective vocabulary. Manual and automatic interpretation reveal features corresponding to uncertainty, exploratory thinking, and self-reflection. Through steering experiments, we provide causal evidence that certain features selected by \score{} directly correspond to the model's reasoning behaviors. Amplifying them prolongs the internal thought process and increases performance on reasoning-related benchmarks. Stage-wise analysis confirms that these features emerge only after reasoning fine-tuning. Our work provides the first mechanistic evidence that specific, interpretable components of LLM representations are causally linked to complex reasoning behaviors.

\section*{Limitations}

\paragraph{ReasonScore.}

Our metric depends on hyperparameters (window size, entropy penalty $\alpha$) that require further ablation studies. Of the $200$ candidates, we found $46$ interpretable features. Although other features can also contribute to reasoning, we could not confidently classify them due to ambiguous activation patterns. Finally, our reasoning vocabulary may not capture all reasoning patterns. These limitations suggest opportunities for future work.

\paragraph{SAEs.}

SAEs provide a powerful interpretability framework. However, it suffers from problems that complicate the extraction of fully interpretable features \cite{chanin2024absorption, leask2025canonical}. This may cause us to miss some features.

\paragraph{Emergence of Reasoning Features.}

Although the results in Sec.~\ref{sec:diffing} support our hypothesis, we acknowledge certain limitations of the diffing approach. The cosine similarity threshold ($0.7$) is empirically chosen following the initial work, and may miss similar features if the representation is rotated during one of the fine-tuning stages. Only $60\%$ of the verified features (and $78\%$ of the $\mathcal{F}_\mathcal{R}$ features) appeared in the final stage, likely due to fine-tuning SAE rather than training from scratch. These limitations show that our approach can result in false negative and false positive predictions. However, we believe that our primary finding remains valid even under these limitations.

\bibliography{custom}

\clearpage

%\onecolumn

\appendix
\include{appendix}

\end{document}

%% file: appendix.tex
\section{ReasonScore Details}
\subsection{Reasoning Vocabulary}
\label{sec:appendix_reasoning_vocabulary}

In Fig.~\ref{fig:reasoning_vocabulary}, we show the complete list of words from the \textit{reasoning} vocabulary $\mathcal{R}$ that we obtain in Sec.~\ref{sec:reasoning_space}. For clarity, we list only the lowercase forms without spaces (e.g. ``alternatively''). However, in our implementation, we track multiple forms of each word, including capitalized (``Alternatively'') and space-prefixed variants (`` alternatively'', `` Alternatively''), as the tokenizer can assign different tokens for each of the forms.

Our vocabulary includes several words that require additional clarification. Concretely, we retain ``but'' despite exceeding the frequency threshold, as a strong reasoning indicator, as shown in prior research. Empirical analysis further confirmed its association with reasoning transitions. We include ``let me'' as a phrase rather than ``let'' alone, since model consistently pair these tokens as a single reasoning unit. For ``but'' and ``another'', we use only capitalized forms, as lowercase variants appear predominantly in non-reasoning contexts while capitalized forms mark reasoning transitions.

\begin{figure}[h!]
    \begin{tcolorbox}[left=1.5mm, right=1.5mm, top=1.5mm, bottom=1.5mm]
    \footnotesize
    \texttt{["alternatively", "hmm", "maybe", "wait", "perhaps", "let me", "therefore", "however", "but", "another"]}
    \end{tcolorbox}
    \caption{The complete list of words from the \textit{reasoning} vocabulary $\mathcal{R}$ in the lowercase and without spaces form.}
    \label{fig:reasoning_vocabulary}
\end{figure}

\subsection{Importance of Reasoning Words}
\label{sec:appendix_functional_importance}

To address whether our selected reasoning words truly reflect model's reasoning capabilities or merely act as superficial linguistic markers, we conduct an ablation study that measures the importance of the reasoning vocabulary on model performance. Specifically, during the model generation, we \textbf{ban} words from our reasoning vocabulary so the model can't output them during its thinking process. We then measure the performance of this setup using $\text{pass}@1$ metric on AIME-2024, MATH-500, and GPQA Diamond. To isolate the impact of reasoning words from the general effect of banning any words, we conducted an additional experiment where we banned an equal number of words, with each word \textit{randomly} sampled from the $p_{\text{think}}$ distribution (Sec.~\ref{sec:reasoning_space}). For the \textit{random words} experiment, we repeat it $10$ times with different sets of words and report average $\text{pass}@1$.

\begin{table}[t!]
    \centering
    \resizebox{\linewidth}{!}{
    \begin{tabular}{@{}l *{6}{c} @{}}
    \toprule
    \multirow{3}{*}{\centering\textbf{Word-ban}} & \multicolumn{2}{c}{\multirow{2}{*}{\textbf{AIME 2024}}} & \multicolumn{2}{c}{\multirow{2}{*}{\textbf{MATH-500}}} & \multicolumn{2}{c}{\textbf{GPQA}} \\
    &  &  &  &  & \multicolumn{2}{c}{\textbf{Diamond}} \\
    \cmidrule(lr){2-3} \cmidrule(lr){4-5} \cmidrule{6-7}
    & pass@1 & tokens (K) & pass@1 & tokens (K) & pass@1 & tokens (K) \\
    \midrule
    \textbf{Default} & 53.3 & 12.4 & 93.2 & 3.9 & 50.0 & 7.9 \\
    \midrule
    \textbf{Reason} & \textbf{20.0} & \textbf{9.1} & \textbf{86.6} & \textbf{2.5} & \textbf{41.9} & \textbf{5.7} \\
    \textbf{Random} & 48.6 & 13.3 & 89.4 & 4.2 & 48.5 & 8.9 \\
    \bottomrule
    \end{tabular}
    }
    \caption{Performance and average number of output tokens (K) under word-ban experiments. \textbf{Default}: no banned words; \textbf{Reason}: ban the reasoning vocabulary; \textbf{Random}: ban random words.}
    \label{tab:functional_importance}
\end{table}

The results, presented in Tab.~\ref{tab:functional_importance}, demonstrate that banning \textit{reasoning} words significantly decreases both the performance and the reasoning depth (average number of output tokens). In contrast, banning random words produces no statistically significant differences from the baseline. This provides empirical evidence that these reasoning words play a functional role in the model's reasoning process.

\subsection{ReasonScore Distribution}
\label{sec:appendix_score_distrinution}

\begin{figure}[h!]
\centering
    \includegraphics[width=1.0\linewidth]{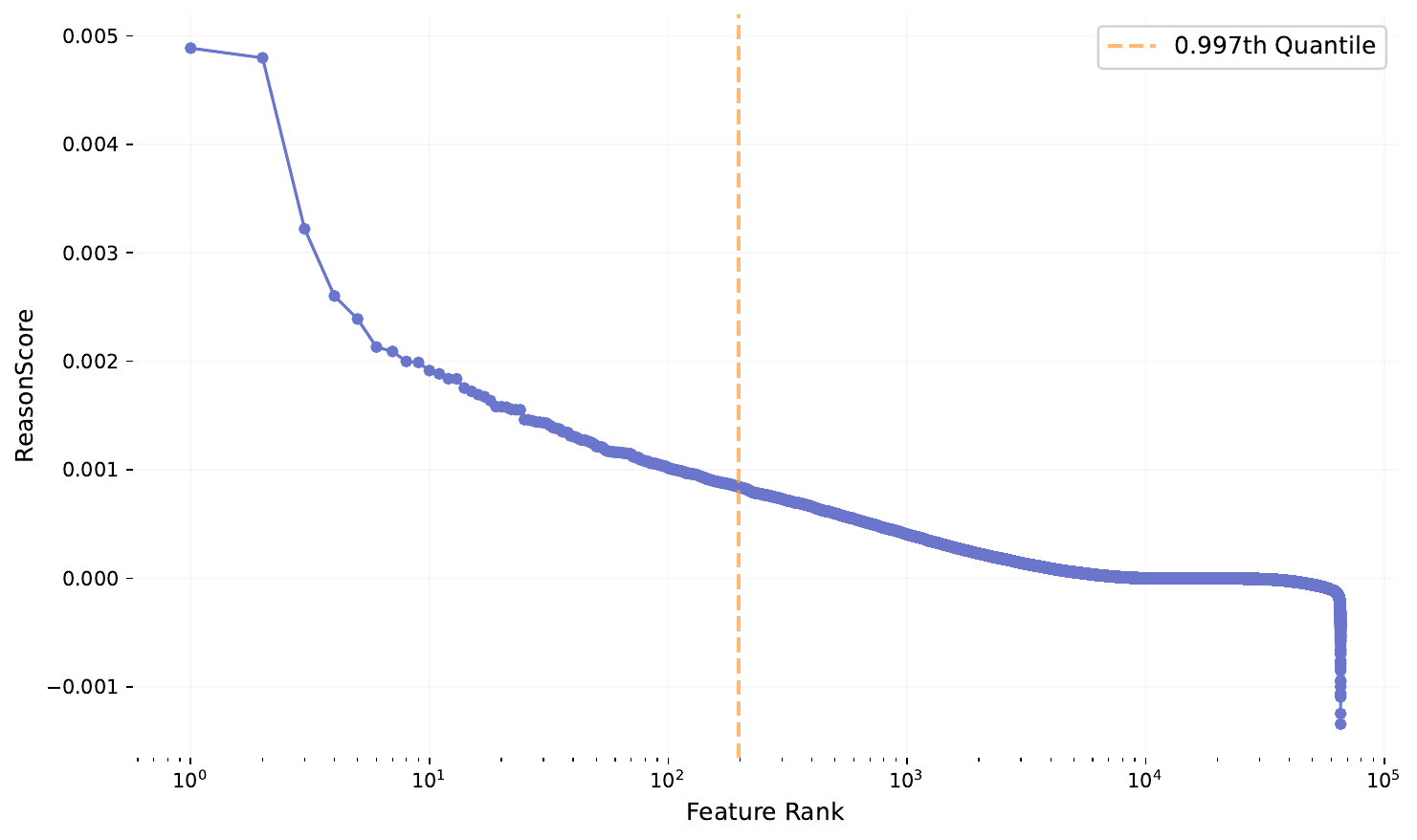}
    \caption{Distribution of ReasonScore values across all SAE features for \textsc{DeepSeek-R1-Llama-8B}, sorted in decreasing order.}
    \label{fig:reasonscore_distribution}
\end{figure}

Fig.~\ref{fig:reasonscore_distribution} shows the ReasonScore values sorted in decreasing order of all SAE features for \textsc{DeepSeek-R1-Llama-8B}. We select the $0.997$-th quantile as a cutoff, yielding approximately $200$ features. While the plot shows the ReasonScore continues to decay below this threshold rather than reaching a plateau, this amount is feasible to analyze manually and contains all the most important features as judged by our metric.

% \newpage

\section{Interpretability Details}
\label{sec:appendix_empirical_analysis}

\subsection{Feature Interfaces}
\label{sec:appendix_feature_interfaces}

\begin{figure*}[t!]
    \centering
    \subfloat[Top-activating examples from the manually verified set of features. The chosen examples represent ``uncertainty''.]
    {
    \label{fig:interface_1}
    \includegraphics[width=0.48\linewidth]{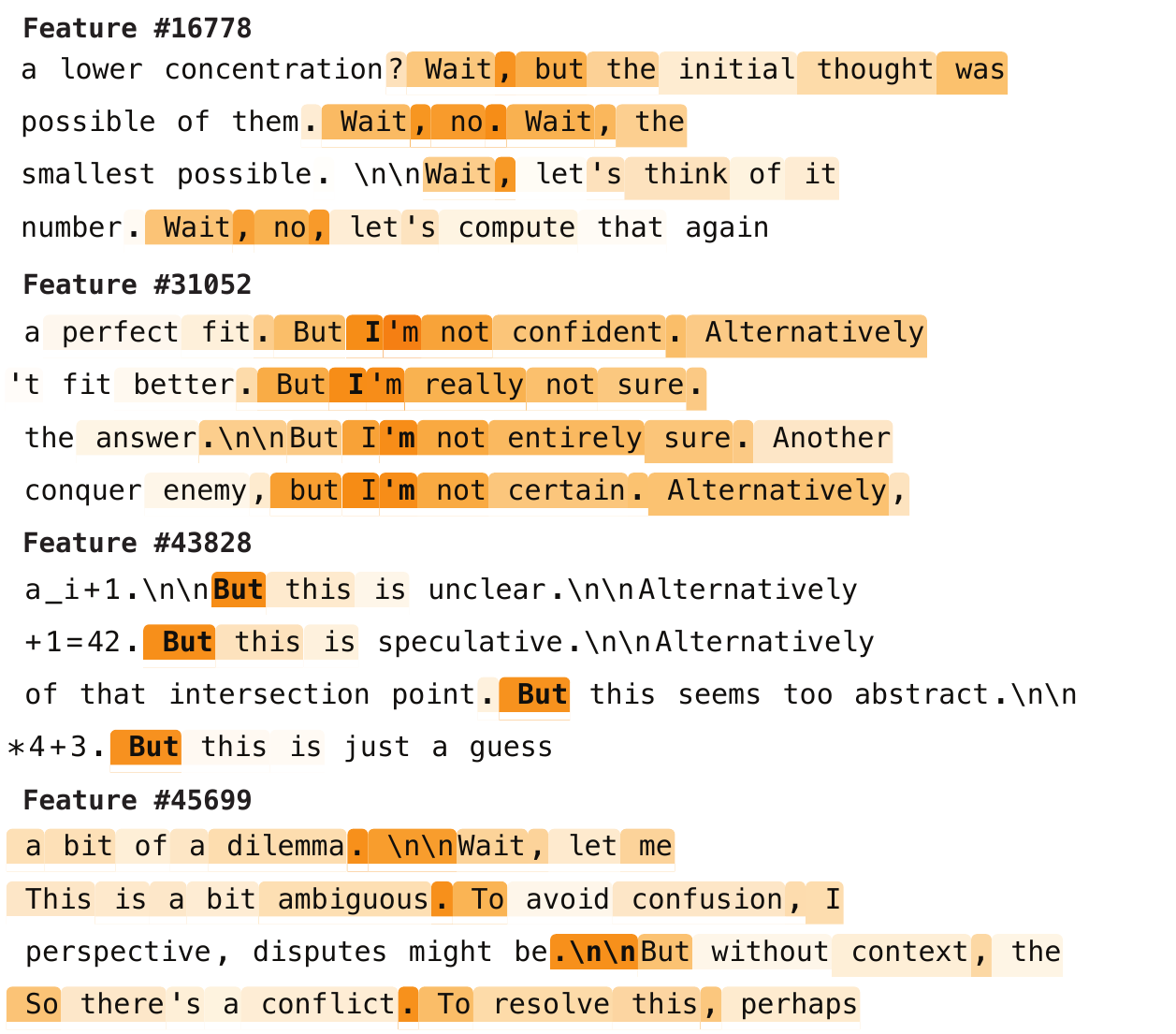}
    }\hfill
    \subfloat[Top-activating examples from the manually verified set of features. The chosen examples represent ``exploration''.]
    {
    \label{fig:interface_2}
    \includegraphics[width=0.48\linewidth]{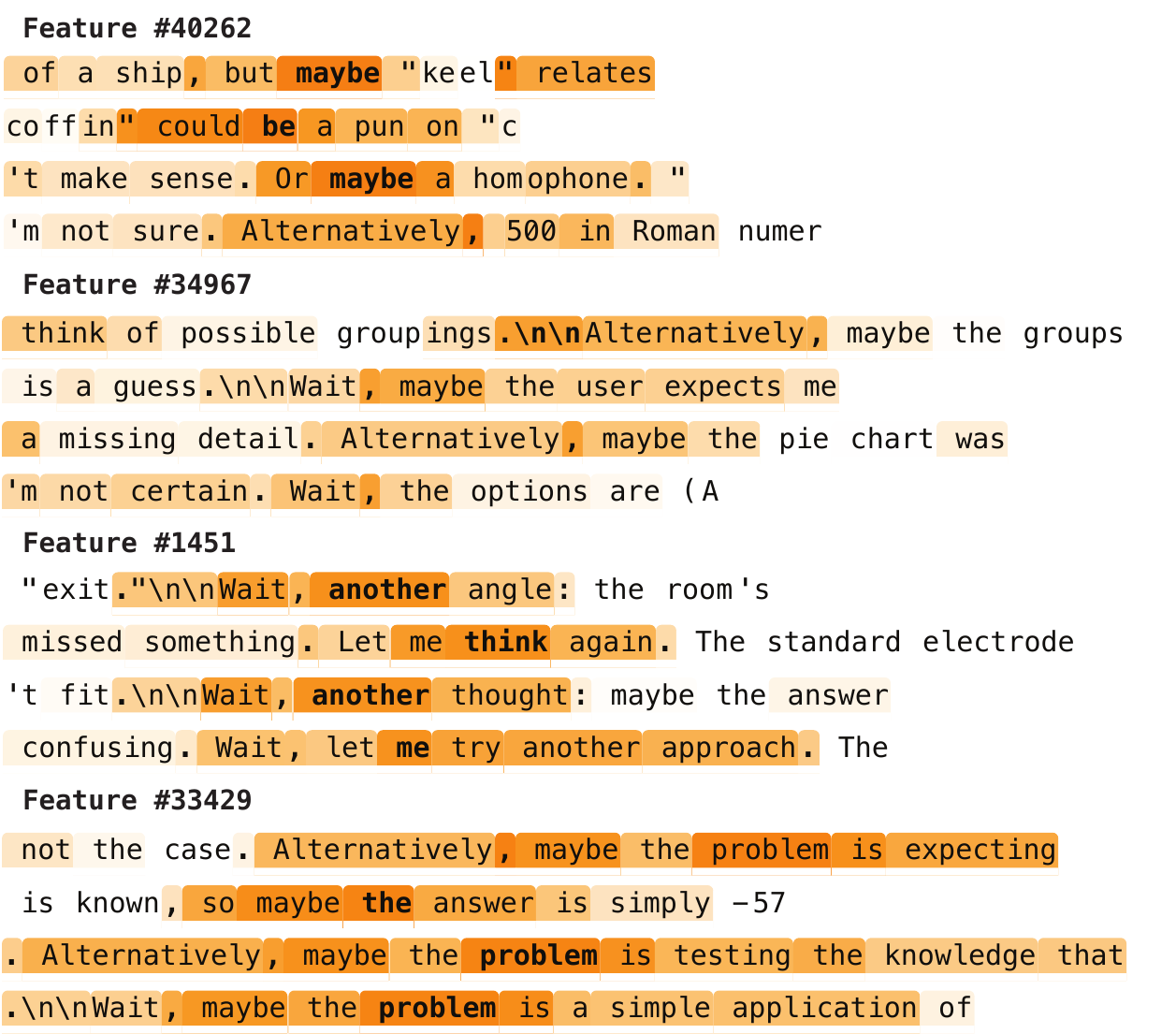}
    }
    \caption{Examples of feature interfaces used in manual interpretation experiments.}
    \label{fig:appendix_interpretation_1}
\end{figure*}

\begin{figure*}[t!]
    \centering
    \subfloat[Top-activating examples from the manually verified set of features. The chosen examples represent ``reflection''.]
    {
    \label{fig:interface_3}
    \includegraphics[width=0.48\linewidth]{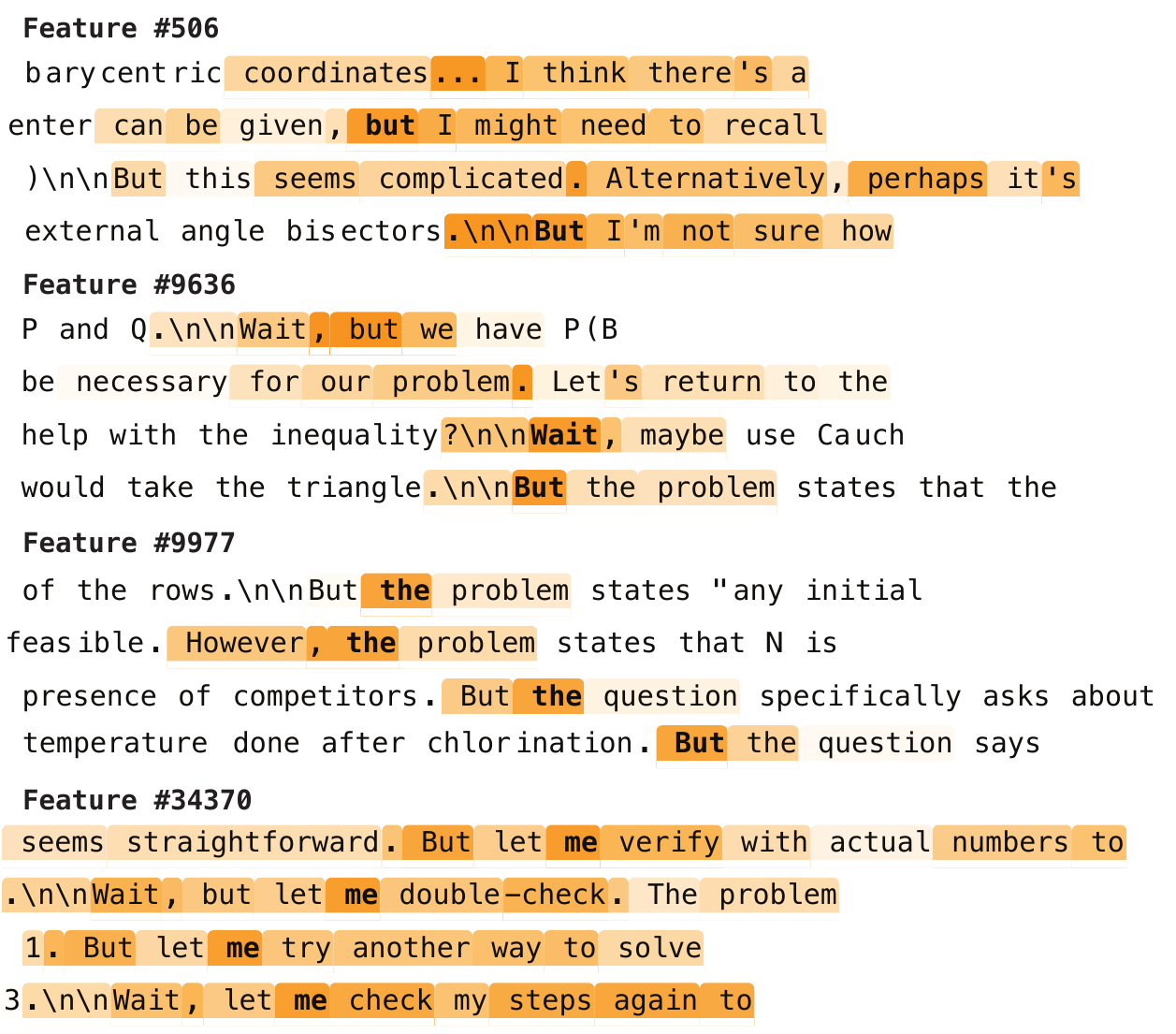}
    }\hfill
    \subfloat[Top-activating examples from the manually verified set of features. The chosen examples represent mixed behaviors.]
    {
    \label{fig:interface_4}
    \includegraphics[width=0.48\linewidth]{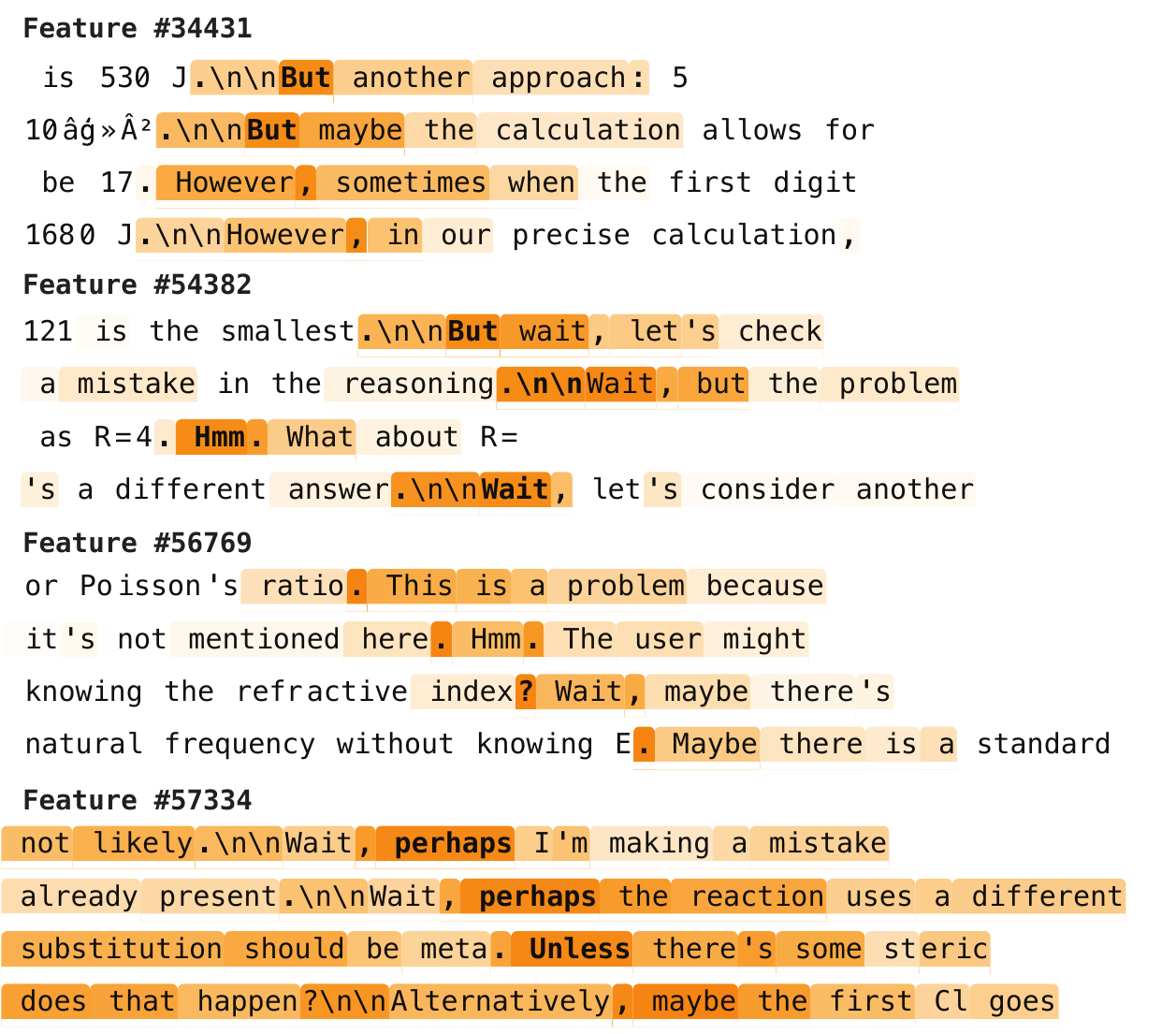}
    }
    \caption{Examples of feature interfaces used in manual interpretation experiments.}
    \label{fig:appendix_interpretation_2}
\end{figure*}

Figs.~\ref{fig:appendix_interpretation_1},\ref{fig:appendix_interpretation_2} display additional activation patterns for features we found during manual interpretation (Sec.~\ref{sec:interpretability}), highlighting tokens where each feature most strongly activates.

\subsection{Automatic Interpretation Details}
\label{sec:appendix_autointerp}

To interpret each feature, we randomly selected 30 examples from the \textsc{OpenThoughts-114k} dataset. The selection was stratified across different domains to ensure a balanced representation of various reasoning types and application contexts. The automatic feature interpretation followed a two-step pipeline. First, we used \textsc{GPT-4o} to generate free-form descriptions of the inferred function of each feature. This step involved detailed prompting, requesting the model to explain what the feature likely encodes and provide supporting examples.

After generating individual descriptions, we cluster reasoning-related features based on their possible functions and behavioral patterns. To support this process, we provided \textsc{GPT-4o} with a list of existing features, accompanied by a description of its possible function and observed steering behavior. The model was asked to identify recurring patterns and group similar features accordingly. To ensure accuracy, the results were then manually reviewed and validated. Table \ref{tab:autointerp_clusters} presents the resulting feature groups, including categories (reasoning depth and thoroughness, self-correction and backtracking, and others), along with descriptions of their roles and effects. In some cases, even features grouped together based on shared function exhibited subtle differences in how they influenced responses; for instance, among features encouraging structural organization, one may focus on logical flow and paragraphing, while another influences transitions between argument steps. Additionally, features often demonstrated overlapping effects with other groups or influenced aspects beyond reasoning alone. For example, affecting the stylistic tone or structure of the output. This suggests that certain features may play a broader role across different types of reasoning and expression, rather than being confined to a single function.

\section{Examples of Feature Steering}
\label{sec:appendix_steering}

Tab.~\ref{tab:intro_steering_full} show the example of model's thinking process on a ``how many r's in the word strawberry'' question with and without steering. Tabs.~\ref{tab:steering_1},\ref{tab:steering_2},\ref{tab:steering_3} show the examples of model's thinking processes on reasoning-related benchmarks with and without steering.

\section{Generalization Experiments}
\label{sec:appendix_generalization}

\subsection{Reasoning Capabilities Across Layers}

\begin{table}[t!]
    \centering
    \resizebox{\linewidth}{!}{
    \begin{tabular}{@{}c *{6}{c} @{}}
    \toprule
    \multirow{2}{*}{\textbf{Layer}} & \multicolumn{3}{c}{\textbf{Reasoning Features}} & \multicolumn{3}{c}{\textbf{Random Features}} \\
    \cmidrule(lr){2-4} \cmidrule(lr){5-7}
    & \textbf{diff} & \textbf{maybe} & \textbf{same} & \textbf{diff} & \textbf{maybe} & \textbf{same} \\
    \midrule
    13 & 52.5 & 14.0 & \textbf{33.5} & 76.0 & 9.0 & 15.0 \\
    15 & 43.5 & 17.0 & \textbf{39.5} & 63.0 & 13.0 & 24.0 \\
    17 & 27.0 & 17.0 & \textbf{56.0} & 49.0 & 13.0 & 38.0 \\
    21 & 17.5 & 23.5 & \textbf{59.0} & 50.0 & 9.0 & 41.0 \\
    23 & 17.5 & 22.0 & \textbf{60.5} & 54.0 & 15.0 & 31.0 \\
    \bottomrule
    \end{tabular}
    }
    \caption{Cross-layer matching results for reasoning features and $100$ random features from layer $19$-th SAE. All values are in percentages ($\%$).}
    \label{tab:matching_results}
\end{table}

We perform additional experiments by analyzing the existence of similar \textit{reasoning} features across layers. We train multiple SAEs on activations from layers $\{13, 15, 17, 21, 23\}$ of the \textsc{DeepSeek-R1-Llama-8B} model. To match the features between different SAEs, we use the approach proposed in \cite{balagansky2024mechanistic}. For each layer, we match all features from our original $19$-th layer SAE with all features in the considered layer. Formally, we aim to find a permutation matrix that aligns the features of layer $A$ with those of layer $B$. This is done by minimizing the MSE between the decoder weights:
\begin{equation}
\begin{aligned}
P^{(A \to B)} &= \arg \min_P \sum_{i=1}^n ||W_{\text{dec},i}^{(A)} - PW_{\text{dec},i}^{(B)}||^2 \\
&= \arg \max_P \langle P, (W_{\text{dec}}^{(A)})^T W_{\text{dec}}^{(B)} \rangle_F.
\end{aligned}
\end{equation}

After matching each $i$-th feature from layer $19$ with some other $j$-th SAE feature from a different layer, we take the subset corresponding to \textit{reasoning} features ($\mathcal{F}_\mathcal{R}$) and their corresponding matched subset. We evaluate the similarity of a pair of matched features $(i, j)$ using a \textit{Matching Score}, which is defined as the cosine similarity between feature activations across a batch of samples from \textsc{OpenThoughts-114k}:
\begin{equation}
\text{MS}(i,j) = \frac{\sum_{x \in D} f_i(x) \cdot f_j(x)}{\sqrt{\sum_{x \in D} f_i^2(x)} \cdot \sqrt{\sum_{x \in D} f_j^2(x)}}.
\end{equation}
Here, $D$ is a subset of token activations from \textsc{OpenThoughts-114k}, $|D| = 10\text{M}$ tokens.

We consider $i$-th reasoning feature to be presented in another layer as $j$-th feature if $\text{MS}(i, j) > 0.7$ (\textbf{same}), undefined if $0.5 < \text{MS}(i, j) \leq 0.7$ (\textbf{maybe}), and absent if $\text{MS}(i, j) \leq 0.5$ (\textbf{diff}).

We present the results in Tab.~\ref{tab:matching_results}, along with the matching scores for the $100$ randomly sampled features from $19$-th layer SAE to establish a baseline for comparison. Our analysis reveals that similar reasoning features exist across model layers (as judged by the \textit{Matching Score}). For example, $39.5\%$ of the features from layer $19$ have close matches ($\text{MS} > 0.7$) in layer $15$, significantly higher than the $24.0\%$ matching rate for randomly chosen features.

\subsection{Reasoning Capabilities Across Architectures}

We train the SAE on the $19$-th layer of a widely adopted open-source family of reasoning models, Nemotron~\cite{bercovich2025llama}. Specifically, we use the \textsc{Llama-3.1-Nemotron-Nano-8B-v1} model and utilize its original training dataset to train our SAE. Otherwise, we follow the same experimental setup as in Sec.~4.1 of the main paper. We compute the \score{} for all features and select the top-200 as judged by the \score{}. 

\begin{figure}[t!]
\centering
    \includegraphics[width=1.0\linewidth]{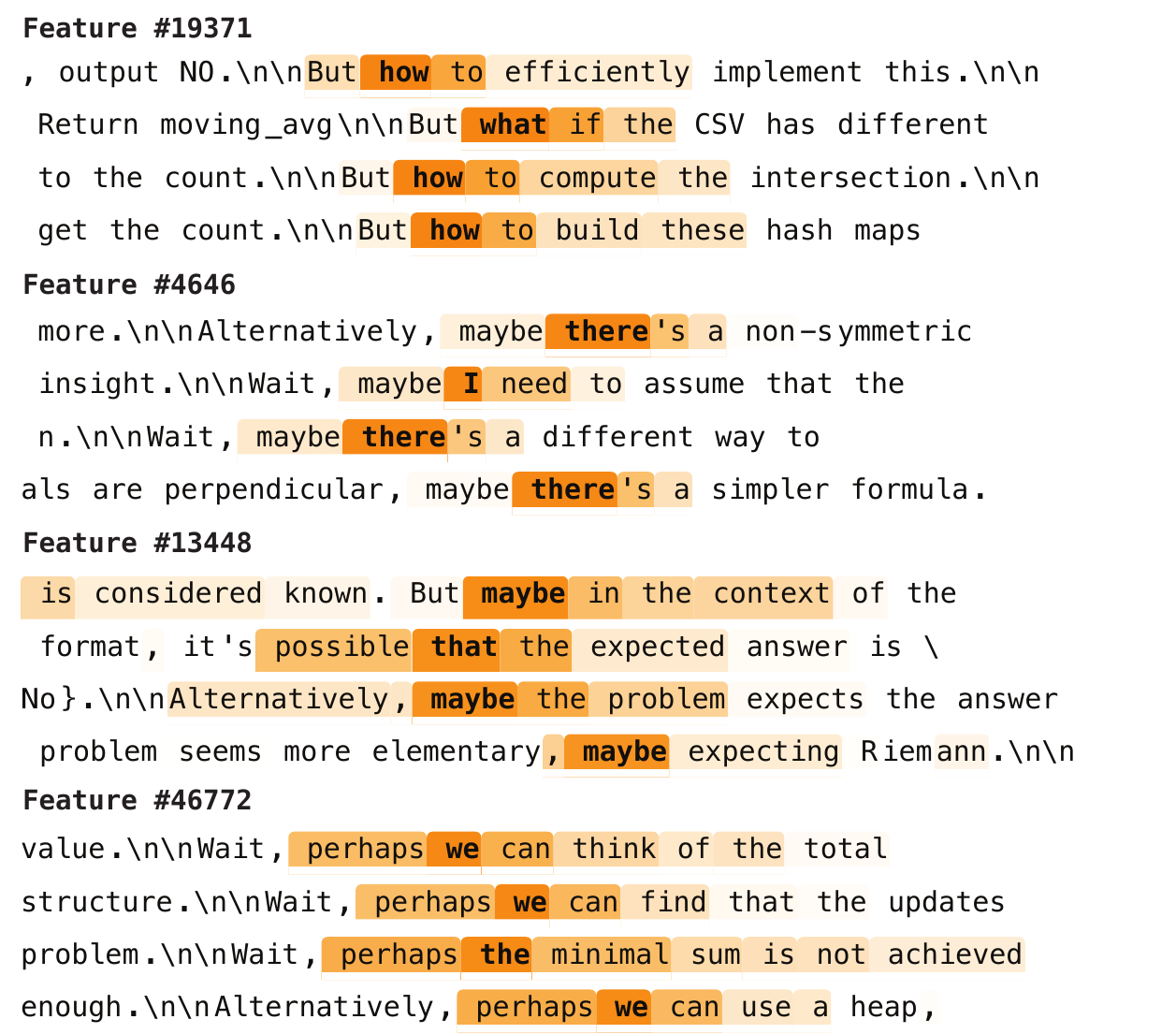}
    \caption{Top-activating examples from the manually verified set of Nemotron features.}
    \label{fig:nemotron_interpretation}
\end{figure}

Following Sec.~4.2 of the main paper, we perform a manual interpretation of these features. We find the Nemotron features that have activation patterns similar to the ones in the \textsc{Deepseek-R1-Llama-8B} model. In Fig.~\ref{fig:nemotron_interpretation}, we provide interfaces of features representing model's uncertainty ($\#46772$), exploration ($\#13448$), and reflection ($\#4646$, $\#19371$). These results indicate that reasoning features emerge across distinct families of reasoning LLMs.

\begin{table*}[h]
    \centering
    \scalebox{0.75}{
    \begin{tabular}{p{.15\textwidth}|p{.15\textwidth}|p{.2\textwidth}|p{.15\textwidth}|p{.3\textwidth}}
    \toprule
        \textbf{Group Name }& \textbf{Features} & \textbf{Possible Function} & \textbf{Effect Type} & \textbf{Observed Behavior} \\ \midrule
        Reasoning Depth and Thoroughness & 506, 4395, 9636, 23052, 30288, 33148, 54382, 61935 & Controls multi-step analysis, iteration, and self-correction in problem-solving. & Stylistic \& Structural, Semantic \& Logical Consistency & \textbf{Strengthening}: Extensive step-by-step reasoning, multiple iterations, self-corrections. \textbf{Weakening}: Direct answers with minimal steps. \\ \hline
        Numerical Accuracy and Validation & 4990, 3466, 16778, 46379, 34813, 51765 & Governs precision in calculations, unit conversions, and error-checking. & Semantic \& Logical Consistency & \textbf{Strengthening}: Meticulous unit tracking, iterative re-evaluation. \textbf{Weakening}: Direct results with potential errors. \\ \hline
        Exploration of Multiple Methods & 22708, 62446 & Encourages evaluating alternative approaches before finalizing solutions. & Semantic \& Logical Exploration & \textbf{Strengthening}: Compares multiple strategies (e.g., DP vs. greedy). \textbf{Weakening}: Commits to the first viable method. \\ \hline
        Structural and Logical Organization & 57334, 43828, 45699, 49326, 17726, 46449, 41636, 40262 & Ensures clarity, step-by-step breakdown, and logical flow. It may also balance executable code generation vs. verbal explanations. & Structural, Semantic \& Instruction Clarity & \textbf{Strengthening}: Well-structured explanations. \textbf{Weakening}: Disorganized or fragmented reasoning. \\ \hline
        Symbolic vs. Numerical Reasoning & 48026, 34967, 34589 & Balances theoretical/symbolic reasoning with direct numerical computation. & Semantic \& Logical Consistency & \textbf{Strengthening}: Algebraic/theoretical frameworks. \textbf{Weakening}: Immediate numerical substitution. \\ \hline
        Self-Correction and Backtracking & 16778, 35337, 42609, 34431, 25953 & Controls iterative refinement and error-checking. & Semantic \& Logical Consistency & \textbf{Strengthening}: Multiple rounds of self-correction. \textbf{Weakening}: Commits to initial answers without revision. \\ \hline
        Causal Chaining \& Scientific Context & 56769, 34370, 3261, 13457 &	Enforces clear multi‑step causal linkages in science/environmental scenarios, modulates temporal reasoning, hypothetical alternatives and scenario simulation &	Semantic (Causality) & \textbf{Strengthening}: yields explicit causal chains, regulates contrastive reasoning, gives clearer timeline-based reasoning. \textbf{Weakening} results in loosely linked assertions or missing intermediate steps, omit historical or causal context
        \\ \hline
        Edge Case and Constraint Handling & 16343, 46691, 3942 & Ensures validation of edge cases and constraints. & Semantic \& Logical Consistency & \textbf{Strengthening}: Explicitly addresses edge cases. \textbf{Weakening}: Assumes valid inputs without verification. \\ \hline
        Semantic Elaboration \& Conceptual Depth & 1451, 33429, 61104, 25485, 45981, 31052, 16441, 53560 &	Shapes depth of domain‑specific explanations, analogies, trade‑off discussions, and interdisciplinary links &	Semantic (Elaboration \& Breadth)& \textbf{Strengthening}: Adds conceptual depth through analogies, trade-offs, and multi-factor explanations. \textbf{Weakening}: Reduces to simple, surface-level or single-cause statements with minimal abstraction. \\ \hline
        Other Behavior & 48792, 9977, 20046, 12036, 32134 & Include: influences engagement and conversational tone, assertiveness/redundancy/structure in text and terminology & Stylistic \& Structural & \textbf{Strengthening:} Creates a more formal, robotic style with rigid structure and a high degree of confidence in statements. \textbf{Weakening:} Makes the style livelier and more conversational, with informal delivery, varied structure, and a moderate level of confidence that includes elements of uncertainty and flexibility.\\ \hline
        \bottomrule
    \end{tabular}
    }
    \caption{Reasoning clusters obtained using \textsc{GPT-4o}. Each cluster is defined by a particular type of reasoning (depth of analysis, numerical checking, code generation), the specific features involved, their hypothesized function in shaping the models' output style and logic, and the observed behaviors that emerge when using the feature steering.}
    \label{tab:autointerp_clusters}
\end{table*}

\begin{table*}[h!]
\centering

\begin{promptbox}{
How many r's are in the word strawberry?
}{
I need to determine how many times the letter "r" appears in the word "strawberry." First, I'll write down the word and break it into individual letters: S, T, R, A, W, B, E, R, R, Y. Next, I'll identify which of these letters are "r": the third, eighth, and ninth letters. Finally, I'll count these "r"s, which gives me a total of $\boxed{three}$.
}{
Okay, so I need to figure out how many r's are in the word "strawberry." Let me think about this step by step. First, I should probably write down the word and break it down into its individual letters. The word is spelled S-T-R-A-W-B-E-R-R-Y. Wait, let me make sure I have that right. Sometimes I get confused with the spelling. I think the correct spelling is "strawberry." So, let me write it out again: S, T, R, A, W, B, E, R, R, Y. Hmm, now that I write it, I see there are two R's after the E. Let me count the letters again to make sure I don't miss any. S (1), T (2), R (3), A (4), W (5), B (6), E (7), R (8), R (9), Y (10). So the word has 10 letters in total. Now, focusing on the R's. The third letter is R, that's one. Then after E, which is the seventh letter, there are two more R's: the eighth and ninth letters. So that's two more. Let me double-check the spelling again to make sure I'm not miscounting. Sometimes I might confuse the letters. I think it's S-T-R-A-W-B-E-R-R-Y, so yes, the two R's are towards the end. Wait, is there another R in the word? Let me think. The word is "strawberry." I think it's spelled with an R after the E, and another R after that. So that's two R's. I don't recall any other R's in the word. Let me go through each letter again: S, T, R, A, W, B, E, R, R, Y. So, yes, the R's are the third, eighth, and ninth letters. Wait, that's three R's. Did I miscount earlier? Wait, no, I think I made a mistake. Let me recount the letters with their positions. Let's write it out: \\
1. S \\
2. T \\
3. R \\
4. A \\
5. W \\
6. B \\
7. E \\ 
8. R \\
9. R \\
10. Y \\
So, the R's are at positions 3, 8, and 9. That makes three R's in total. I think I initially thought it was two, but upon a closer examination, it's actually three. Let me check another source or think about how the word is pronounced. The word "strawberry" is pronounced with a double R sound, so that would correspond to two R's, but when I break it down, I see that the letters are R, R. Wait, no, the letters are E, R, R. So, that's two R's after the E. So, the third letter is R, and then the eighth and ninth letters are R's as well. So that's three R's. Wait, I'm getting confused. Let me visualize the word again: S-T-R-A-W-B-E-R-R-Y. So, the letters are S, T, R, A, W, B, E, R, R, Y. So, the R appears at the third position, and then again at the eighth and ninth positions. That makes three R's. So, the correct count is three R's. I think I initially miscounted because I focused on the double R at the end, but there's another R earlier in the word. So, to clarify, the word "strawberry" has three R's: one in the third position, and two more at the end. So, the correct answer is $\boxed{three}$ R's.
}
\end{promptbox}

\caption{An example of ``How many r's are in the word strawberry?'' problem and corresponding full outputs from \textsc{DeepSeek-R1-Llama-8B} and its steered version.}
\label{tab:intro_steering_full}
\end{table*}

\begin{table*}[h]
\centering
\begin{promptbox}{
Tom got a Mr. Potato Head for his birthday. It came with 3 hairstyles, 2 sets of eyebrows, 1 pair of googly eyes, 2 sets of ears, and 2 sets of lips, a pair of regular shoes, and a bonus pair of clown shoes. If a complete Mr. Potato Head personality includes eyebrows, eyes, ears, lips, shoes and optionally hair, how many different wacky personalities can Tom come up with? Note that Mr. Potato Head can be bald. Note: You cannot "mix and match".  For example, you cannot take the left eyebrow from one pair and the right eyebrow from the other pair.}{
First, I need to understand the components that make up a complete Mr. Potato Head personality. According to the problem, a personality includes eyebrows, eyes, ears, lips, shoes, and optionally hair. Additionally, Tom received various options for each of these components. Starting with the eyes, there is only one pair of googly eyes available. Since they must be used as a whole pair, there is only 1 way to choose the eyes. Next, the ears come in 2 sets. Each set represents a different style, so there are 2 possible choices for the ears. For the eyebrows, there are 2 sets available. Each set is used entirely, so there are 2 choices for the eyebrows. The lips also come in 2 sets, meaning there are 2 options for the lips. Shoes are either regular or clown shoes, and since both are complete pairs, there are 2 choices for the shoes. Hair is optional and includes 3 different hairstyles. Therefore, there are 3 choices for the hair, but it's not required to choose any. \\
Now, to calculate the total number of different personalities, I'll multiply the number of choices for each component. The eyes, ears, eyebrows, and lips each have 1, 2, 2, and 2 choices respectively. The shoes have 2 choices, and the optional hair adds 3 possibilities. \\
So, the total number of personalities is 1 (eyes) × 2 (ears) × 2 (eyebrows) × 2 (lips) × 2 (shoes) × 3 (hair) = $\boxed{24}$ different personalities.
}{
Alright, so Tom got this Mr. Potato Head, and he wants to figure out how many different wacky personalities he can create. The problem gives a list of parts: 3 hairstyles, 2 sets of eyebrows, 1 pair of googly eyes, 2 sets of ears, 2 sets of lips, a pair of regular shoes, and a bonus pair of clown shoes. The key here is that a complete personality includes eyebrows, eyes, ears, lips, shoes, and optionally hair. Also, it's important to note that you can't mix and match parts; you have to use the entire pair or set as given. First, let me break down the components and see how they contribute to the total number of personalities. The problem mentions that some parts are optional, specifically the hairstyles. So, the personality can be bald, meaning the hairstyle part is optional. Let's list out all the components: \\
1. Hairstyles: 3 options. These can be considered optional since the problem states that Mr. Potato Head can be bald. \\
2. Eyebrows: 2 sets. Since both eyes and eyebrows are required, we have to choose one set of eyebrows. So, there are 2 choices here. \\
3. Eyes: 1 pair of googly eyes. It seems like there's only one option here because it's just a single pair. So, eyes are required, and there's only one choice for them. \\
4. Ears: 2 sets. Similar to eyebrows, we have to choose one set of ears, so that's 2 choices. \\
5. Lips: 2 sets. Again, choosing one set, so 2 choices. \\
6. Shoes: 2 types (regular and clown). Since shoes are required, we have two options here. \\
Now, let's structure this. The components that are required are eyes, eyebrows, ears, lips, and shoes. The optional component is the hairstyle. Since the hairstyles can be chosen or not, that adds some flexibility. So, let's calculate the number of options for each required component: Eyebrows: 2 options, Eyes: 1 option, Ears: 2 options, Lips: 2 options, Shoes: 2 options. Multiplying these together gives us the number of ways to choose the required components: 2 * 1 * 2 * 2 * 2. Calculating that: 2 * 1 is 2, times 2 is 4, times 2 is 8, times 2 is 16. So, there are 16 ways to choose the required components. \\
Now, considering the optional hairstyle. Since there are 3 hairstyles and the Mr. Potato Head can be bald, that means for each combination of the required parts, there are 4 possible choices (including the option of no hairstyle). Wait, hold on, actually, the problem says that the complete personality includes eyebrows, eyes, ears, lips, shoes, and optionally hair. So, does that mean that the hair is optional on top of the required parts? So, for each required component setup, we can choose one of the 3 hairstyles or choose to be bald. Hmm, does being bald count as a separate option, or is it just not using any hairstyle? Wait, the problem says "you cannot 'mix and match'", but that's about using entire sets, not about the hairstyle. So, if the problem allows being bald, that's a fourth option for the hairstyle. Therefore, the number of choices for the hairstyle is 4. \\
So, now, the total number of personalities would be the number of ways to choose the required components multiplied by the number of choices for the hairstyle. Earlier, we found that the required components can be arranged in 16 ways. For each of these 16 ways, we can choose one of the 4 hairstyles (including bald). So, that would be 16 * 4. Calculating that: 16 * 4 is 64. \\
But let me think again. Is the pair of regular shoes considered as 1 option or 2? Wait, the problem says a pair of regular shoes and a bonus pair of clown shoes. So, the shoes are either regular or clown. So, that's two options, right? So, that's 2 choices for shoes. So, that seems correct. So, summarizing: Eyebrows: 2, Eyes: 1, Ears: 2, Lips: 2, Shoes: 2. Multiply those: 2*1*2*2*2 = 16. Hair: 3 + 1 (bald) = 4. Total: 16*4 = 64. Therefore, the total number of different wacky personalities Tom can come up with is 64. \\
Wait, just to make sure, is there any possibility that the problem is considering the shoes as a separate category? Let me check the problem again. It says: "a pair of regular shoes, and a bonus pair of clown shoes." So, that's two different shoes: regular and clown. So, the shoes are either regular or clown, so two options. Therefore, the shoes contribute 2 to the total, which we included. So, the calculations are correct. Therefore, the final answer is $\boxed{64}$.
}
\end{promptbox}
\caption{
A problem from \textsc{MATH-500}, and corresponding outputs from \textsc{DeepSeek-R1-Llama-8B} and its steered version. Correct answer: 64.
}
\label{tab:steering_1}
\end{table*}

% \newpage

\begin{table*}[h]
\centering
\begin{promptbox}{
A regular octagon has the same perimeter as the regular hexagon shown here with side length 16 cm.  How long is each side of the octagon? [asy]size(80); pair A = dir(120); pair B=dir(60); pair M=(A+B)/2; draw(dir(360)--B--A--dir(180)--dir(240)--dir(300)--cycle); label("16 cm", M, N);[/asy]
}{
Okay, so I have this problem where a regular octagon has the same perimeter as a regular hexagon. The hexagon has a side length of 16 cm, and I need to find the side length of the octagon. Hmm, let's break this down step by step. First, I remember that the perimeter of a regular polygon is just the number of sides multiplied by the length of each side. So, for the hexagon, which is a six-sided polygon, I can calculate its perimeter by multiplying 6 by 16 cm. Let me write that out: \\
Perimeter of hexagon = 6 * 16 cm = 96 cm. \\
Got that. So the perimeter of the octagon must also be 96 cm since it's equal to the hexagon's perimeter. Now, a regular octagon has 8 sides, right? So, if I let 's' be the length of each side of the octagon, then the perimeter of the octagon would be 8 * s. Since we know the perimeter is 96 cm, I can set up the equation: \\
8 * s = 96 cm. \\
To find 's', I just need to divide both sides by 8: \\
s = 96 cm / 8 = 12 cm. \\
Wait, is that it? It seems straightforward, but let me double-check to make sure I didn't miss anything. So, the hexagon has 6 sides, each 16 cm, so 6*16=96 cm. The octagon has 8 sides, so each side is 96/8=12 cm. Yep, that seems correct. Hmm, just to make sure, maybe I should visualize the polygons. A regular hexagon has six equal sides and six equal angles, and a regular octagon has eight equal sides and eight equal angles. Since both are regular, all sides and angles are equal, so the calculation should hold. Another way to think about it: the perimeter is just the sum of all sides. So, for the hexagon, adding up six sides each of 16 cm gives 96 cm. For the octagon, we need to add up eight sides, each of which is the same length, so 8 times that length equals 96 cm. Solving for that length gives 12 cm. I don't see any mistakes in the logic here. It's a direct calculation based on the definition of perimeter for regular polygons. So, I think the side length of the octagon is 12 cm. \\ **Final Answer** \\ The length of each side of the octagon is $\boxed{12}$ centimeters.
}{
Okay, so I have this problem here where I need to figure out the length of each side of a regular octagon. The octagon has the same perimeter as a regular hexagon, and the hexagon is shown with side length 16 cm. Hmm, okay, let me think step by step about how to approach this. First, I remember that the perimeter of a shape is the total length around it. For regular polygons, which both octagons and hexagons are, the perimeter is just the number of sides multiplied by the length of each side. So if I can figure out how many sides each shape has and then use the given information, I can find the opposite. Wait, the problem says it's a regular hexagon. Right, a hexagon has six sides. So, since the side length is 16 cm, the perimeter of the hexagon would be 6 times 16 cm. Let me write that down. \\
Perimeter of hexagon = 6 * 16 cm = 96 cm. \\
So, the octagon has the same perimeter, which is 96 cm. Now, an octagon has eight sides. So, if I can find the length of each side of the octagon, that would be the perimeter divided by 8. Let me write that formula: \\
Perimeter of octagon = 8 * (length of each side) \\
But wait, we know the perimeter is 96 cm, so: \\
96 cm = 8 * (length of each side) \\
To find the length of each side, I need to divide both sides by 8. So, length of each side = 96 cm / 8 = 12 cm. Hmm, that seems straightforward. But let me just double-check to make sure I didn't make a mistake. Wait, is it correct that the octagon's perimeter is 96 cm? Because the hexagon's side is 16 cm, so 6 times 16 is 96. So, yes, the octagon has to have the same perimeter. Since octagons have more sides, each side should be shorter. So, 96 divided by 8 is 12, which is shorter than 16, so that makes sense. Wait, another way to think about it: if you have a polygon with more sides, each side is actually shorter for the same perimeter. So, since an octagon has more sides than a hexagon, each side should be shorter, which aligns with 12 cm being shorter than 16 cm. So that seems right. Is there another way to approach this problem? Maybe using the formula for perimeter in terms of side length? Yes, for regular polygons, the perimeter is just the number of sides multiplied by the side length. So, if I denote the side length of the octagon as 's', then: \\
Perimeter of octagon = 8 * s \\
But since it's equal to the perimeter of the hexagon, which is 6 * 16 = 96, then: 8s = 96 Solving for s gives s = 96 / 8 = 12 cm. Yep, same result. I think I got it. So, the length of each side of the octagon is 12 cm. Wait, let me think again. Sometimes, when I'm dealing with perimeters, I might confuse side length with something else, but in this case, since both are regular polygons, it's straightforward. So, for regular polygons, the perimeter is directly additive. So, each side contributes equally. Therefore, inverting the number of sides inversely affects the side length. So, in the hexagon, 6 sides, each 16 cm, so 16 times 6 is 96. The octagon, which has more sides, 8, so each side is 96 divided by 8, which is 12. So, that makes sense. Wait, just to make sure, if the octagon had 8 sides, each 12 cm, the perimeter would be 8*12=96 cm, which is the same as the hexagon. So, that's correct. I think that's solid. So, the length of each side of the octagon is 12 cm. \\ **Final Answer** \\ The length of each side of the octagon is $\boxed{12}$ cm.
}
\end{promptbox}
\caption{
A problem from \textsc{MATH-500}, and corresponding outputs from \textsc{DeepSeek-R1-Llama-8B} and its steered version. Correct answer: 12 cm.
}
\label{tab:steering_2}
\end{table*}

% \newpage

\begin{table*}[h]
\centering
\begin{promptbox}{
Remmy wants to divide $10$ by $\frac{2}{3}$, but he cannot remember how to do that.  By what number should he multiply $10$ to get the answer?
}{
Okay, so Remmy has this problem where he needs to divide 10 by 2/3, but he's forgotten how to do division with fractions. Hmm, I remember that dividing by a fraction is the same as multiplying by its reciprocal, but let me think through it step by step to make sure I get it right. First, division. If we have a division problem like 10 divided by something, in this case, 10 divided by 2/3, the basic idea is to find out how many times 2/3 fits into 10. But when dealing with fractions, it can get a bit tricky, so maybe breaking it down will help. I recall that dividing by a fraction is equivalent to multiplying by its reciprocal. The reciprocal of a fraction is just flipping the numerator and denominator. So, the reciprocal of 2/3 is 3/2. That means 10 divided by 2/3 is the same as 10 multiplied by 3/2. Let me write that out: 10 ÷ (2/3) = 10 × (3/2) \\
Okay, so now I just need to compute 10 multiplied by 3/2. To do that, I can think of 10 as a fraction, which is 10/1. Then, multiplying two fractions is straightforward: multiply the numerators and the denominators. So, 10/1 × 3/2 = (10 × 3)/(1 × 2) = 30/2. Hmm, 30 divided by 2 is 15. So, 10 divided by 2/3 is 15. That makes sense because 2/3 is less than 1, so dividing by a number less than 1 should give a result larger than the original number. \\
Wait, let me double-check that. If I have 2/3, how many times does it go into 10? Well, 2/3 times 15 should be 10. Let me verify that: (2/3) × 15 = (2 × 15)/3 = 30/3 = 10. Yes, that's correct. So, 15 is the right answer. \\
Alternatively, maybe I can think about this in another way to confirm. If I have 10 divided by 2/3, it's like asking how much do I have to multiply 2/3 by to get 10. So, if I set it up as an equation: (2/3) × x = 10. To solve for x, I can multiply both sides by the reciprocal of 2/3, which is 3/2: x = 10 × (3/2) = 15. Same result. That seems consistent. \\
Another way to think about division with fractions is by using decimals. 2/3 as a decimal is approximately 0.6667. So, if I divide 10 by 0.6667, what do I get? Let me calculate that. 10 ÷ 0.6667 is approximately 15.0. That matches the previous answers, so that's good. Hmm, so all methods lead me to 15. That must be the correct answer. Just to recap: Dividing by a fraction is the same as multiplying by its reciprocal; The reciprocal of 2/3 is 3/2; 10 × 3/2 = 30/2 = 15; Verification by multiplying 2/3 by 15 gives back 10; Converting to decimals also gives 15. Therefore, I'm confident that the answer is 15. \\
**Final Answer** The number Remmy should multiply 10 by is $\boxed{15}$.
}{
Okay, so Remmy wants to divide 10 by 2/3, but he's a bit fuzzy on how to do that. Hmm, I remember that dividing by a fraction is the same as multiplying by its reciprocal. Let me think if that's right. So, if you have something like 10 divided by 2/3, it should be the same as 10 multiplied by 3/2. Yeah, that sounds familiar from what I've learned before. \\
Wait, let me make sure I'm not mixing things up. Sometimes, when you see division, you might be tempted to think of it as subtracting, but no, division is more about finding how many times one number fits into another. So, when you divide by a fraction, you're actually scaling up the original number because the divisor is less than 1. That makes sense because 2/3 is less than 1, so dividing by it should give a larger result than 10. Let me break it down step by step. The problem is 10 divided by 2/3. To make it clearer, I can write that as 10 ÷ (2/3). Now, dividing by a fraction is equivalent to multiplying by its reciprocal. The reciprocal of 2/3 is 3/2, right? So, if I flip the divisor, it becomes 3/2. So, now the problem becomes 10 multiplied by 3/2. Let me compute that. 10 times 3 is 30, and 30 divided by 2 is 15. So, the result of 10 divided by 2/3 is 15. \\
But wait, the question isn't asking for the answer itself; it's asking by what number Remmy should multiply 10 to get the answer. So, in this case, it's 3/2. Hold on, let me verify that again. If I have 10 and I multiply it by 3/2, I get 15, which is indeed the result of dividing 10 by 2/3. So, yes, 3/2 is the correct multiplier. Alternatively, I can think of it as 10 multiplied by 1.5, which is also 15. So, both ways, it checks out. Is there another way to approach this problem? Maybe by converting the division into a multiplication problem from the start. So, instead of 10 ÷ (2/3), I can rewrite this as 10 multiplied by the reciprocal of 2/3, which is 3/2. That again gives me 15 when I do the multiplication. So, that's consistent. \\
I wonder if I can visualize this. If I have 10 wholes and I want to divide them into portions of 2/3 each, how many portions would I have? Well, each portion is 2/3, so to get 10 portions, I need to figure out how much each portion is. But wait, that's the opposite. Maybe that's confusing. Let me think differently. \\
Suppose I have 10 cookies, and each cookie is cut into 3 equal pieces. Each piece would be 1/3 of a cookie. But if I don't cut them all the way, maybe I cut each cookie into 2/3 pieces. Hmm, not sure if that helps. Alternatively, if I have 10 apples and I want to distribute them among 2/3 of a basket each, how many baskets would I need? But maybe that's overcomplicating it. \\
Perhaps it's better to stick with the mathematical approach. So, 10 divided by 2/3 is the same as 10 times 3/2, which is 15. Therefore, Remmy should multiply 10 by 3/2 to get the answer. \\
Let me also think about the properties of division and multiplication. Division can be seen as the inverse operation of multiplication. So, if you have a ÷ (b/c), it's equivalent to a * (c/b). Applying that here, 10 ÷ (2/3) is 10 * (3/2). Yep, that's exactly what I did earlier. Just to make sure I haven't made any calculation errors, let me do the multiplication again. 10 times 3 is 30, and 30 divided by 2 is 15. So, 10 times 3/2 is 15. That seems right. Alternatively, if I think about decimals, 2/3 is approximately 0.6667. So, 10 divided by 0.6667 should be roughly 15. Let me compute that. 10 divided by 0.6667. Well, 0.6667 goes into 10 about 15 times because 0.6667 times 15 is 10. That's a good check. I think I'm confident now that the number Remmy should multiply 10 by is 3/2. So, 3/2 is the correct multiplier. \\
**Final Answer** Remmy should multiply 10 by $\boxed{3/2}$.
}
\end{promptbox}
\caption{
A problem from \textsc{MATH-500}, and corresponding outputs from \textsc{DeepSeek-R1-Llama-8B} and its steered version. Correct answer: $3/2$.
}
\label{tab:steering_3}
\end{table*}

% \newpage

\begin{table*}[t]
\centering
\begin{promptbox}{
A curve is parameterized by \\ $(x,y) = (t^3 + 7, -3t^2 - 6t - 5).$ \\ Find the point the curve passes through at $t = 2.$
}{
Okay, so I have this problem here where a curve is parameterized by the equations \( x = t^3 + 7 \) and \( y = -3t^2 - 6t - 5 \). I need to find the point the curve passes through when \( t = 2 \). Hmm, let me think about how to approach this. First, I remember that when you have a parameterized curve, each value of \( t \) corresponds to a specific point \( (x, y) \) on the curve. So, to find the point when \( t = 2 \), I just need to substitute \( t = 2 \) into both the equations for \( x \) and \( y \) and compute the values. That makes sense because each parameter \( t \) defines a unique point on the curve. Alright, let's start with the equation for \( x \). The equation is \( x = t^3 + 7 \). So, substituting \( t = 2 \) into this should give me the x-coordinate of the point. Let me compute that step by step. Calculating \( t^3 \) when \( t = 2 \): \( 2^3 = 8 \). Then, adding 7 to that: \( 8 + 7 = 15 \). So, the x-coordinate is 15. That seems straightforward. Now, moving on to the y-coordinate. The equation is \( y = -3t^2 - 6t - 5 \). Again, substituting \( t = 2 \) into this equation. Let me break it down. First, compute \( t^2 \) when \( t = 2 \): \( 2^2 = 4 \). Then, multiply that by -3: \( -3 \times 4 = -12 \). Next, compute \( -6t \) when \( t = 2 \): \( -6 \times 2 = -12 \). Lastly, we have the constant term, which is -5. So, putting it all together: \( y = (-12) + (-12) + (-5) \). Wait, hold on. Let me double-check that. The equation is \( y = -3t^2 - 6t - 5 \). So, substituting t = 2, it's \( -3*(2)^2 -6*(2) -5 \). That is \( -3*4 -12 -5 \). So, -12 -12 -5. Adding those together: -12 -12 is -24, and -24 -5 is -29. So, the y-coordinate is -29. Let me make sure I did that correctly. So, \( -3*(2)^2 \) is \( -3*4 = -12 \). Then, -6*2 is -12, and then subtracting 5. So, adding all those up: -12 -12 is -24, minus another 5 is -29. Yeah, that seems right. So, combining both coordinates, when t = 2, the point on the curve is (15, -29). Hmm, let me just verify once more in case I made any calculation errors. Starting with x: \( t^3 + 7 \). t = 2, so \( 2^3 \) is 8, plus 7 is 15. That's correct. For y: \( -3t^2 - 6t -5 \). t = 2: \( -3*(4) = -12 \), \( -6*2 = -12 \), and the constant is -5. So, adding them: -12 -12 is -24, minus 5 is -29. That's correct. Okay, so I think that's the point. Just to visualize, when t = 2, the curve is at (15, -29). I don't see any mistakes in my calculations, so I feel confident about this answer. \\ **Final Answer** \\ The point the curve passes through at \( t = 2 \) is $\boxed{(15, -29)}$.
}{
Alright, so I have this problem here where a curve is parameterized by the equations \(x = t^3 + 7\) and \(y = -3t^2 - 6t - 5\). I need to find the point the curve passes through when \(t = 2\). Hmm, okay, let me think about how to approach this. First, I remember that parameterized equations give both the x and y coordinates in terms of a third variable, which in this case is \(t\). So, when they ask for the point when \(t = 2\), I need to substitute \(t = 2\) into both equations for \(x\) and \(y\) and then find the corresponding \((x, y)\) pair. Let me write that down step by step. So, for \(x\), the equation is \(x = t^3 + 7\). If I plug in \(t = 2\), that should give me the x-coordinate. Similarly, for \(y\), the equation is \(y = -3t^2 - 6t - 5\), so substituting \(t = 2\) there should give me the y-coordinate. Once I have both values, I can combine them into a point. Okay, let me start with calculating \(x\). Plugging \(t = 2\) into \(x = t^3 + 7\): \\
\(x = (2)^3 + 7\). \\
Calculating \(2^3\), that's \(8\), right? So, \(8 + 7\) is \(15\). So, the x-coordinate when \(t = 2\) is \(15\). That seems straightforward. Now, moving on to the y-coordinate. The equation is \(y = -3t^2 - 6t - 5\). Plugging in \(t = 2\): \\
\(y = -3(2)^2 - 6(2) - 5\). \\
Let me compute each term step by step. First, \(2^2\) is \(4\), so multiplying that by \(-3\) gives \(-12\). Next, \(6 \times 2\) is \(12\), so with the negative sign, that term becomes \(-12\). Then, the last term is just \(-5\). So, putting it all together: \(-12 - 12 - 5\). Let's add those numbers. \(-12 - 12\) is \(-24\), and then subtracting another 5 gives \(-29\). So, the y-coordinate is \(-29\). Wait, let me double-check that calculation because sometimes when I have multiple negative signs, I might mix them up. So, let me recast it: \\
\(y = -3(4) - 6(2) - 5\). \\
That's \(-12 - 12 - 5\). Yes, that's correct. \(-12 - 12\) is indeed \( -24 \), and \(-24 - 5\) is \(-29\). So, that seems right. So, to recap, when \(t = 2\), the x-coordinate is 15 and the y-coordinate is \(-29\). Therefore, the point the curve passes through is \((15, -29)\). Wait, just to make sure I didn't make a mistake in the substitution, let me verify each step again. For \(x\), substituting \(t = 2\) into \(t^3 + 7\): \(2^3 = 8\), so \(8 + 7 = 15\). That's correct. For \(y\), substituting \(t = 2\) into \(-3t^2 - 6t - 5\): \\
First term: \(-3*(2)^2 = -3*4 = -12\). \\
Second term: \(-6*2 = -12\). \\
Third term: \(-5\). \\
Adding them up: \(-12 -12 -5 = (-12 -12) -5 = -24 -5 = -29\). Yep, that's correct. Hmm, I think that's solid. Maybe I should just think about whether the parameterization makes sense or if there's another way to represent this curve. Wait, actually, I might recall that sometimes parameterizations can be tricky, but in this case, it's pretty straightforward. It's giving x and y in terms of t, so plugging in t=2 should give the point. Alternatively, I can also compute the values for t=2 in both equations and plot them mentally or on paper. Let me sketch a rough graph in my mind. So, when t=2, x is 15, which is further out on the x-axis, and y is -29, which is quite low on the y-axis. That seems plausible for a cubic and quadratic parameterization. Wait, another thought: sometimes, when I parameterize, especially with higher powers, I might have to check for consistency, but in this case, since both x and y are given, and the question is just about plugging in a specific t, I think it's safe. So, to sum up, substituting t=2 into both equations gives x=15 and y=-29, so the point is (15, -29). I think that's the correct answer. \\ **Final Answer** \\ The point the curve passes through at $t = 2$ is $\boxed{(15, -29)}$.
}
\end{promptbox}
\caption{
A problem from \textsc{MATH-500}, and corresponding outputs from \textsc{DeepSeek-R1-Llama-8B} and its steered version. Correct answer: (15, -29).
}
\label{tab:steering_4}
\end{table*}

\clearpage

%% file: arxiv.bbl
\begin{thebibliography}{64}
\providecommand{\natexlab}[1]{#1}

\bibitem[{Alain and Bengio(2016)}]{alain2016understanding}
Guillaume Alain and Yoshua Bengio. 2016.
\newblock Understanding intermediate layers using linear classifier probes.
\newblock \emph{arXiv preprint arXiv:1610.01644}.

\bibitem[{Balagansky et~al.(2024)Balagansky, Maksimov, and Gavrilov}]{balagansky2024mechanistic}
Nikita Balagansky, Ian Maksimov, and Daniil Gavrilov. 2024.
\newblock Mechanistic permutability: Match features across layers.
\newblock \emph{arXiv preprint arXiv:2410.07656}.

\bibitem[{Bercovich et~al.(2025)Bercovich, Levy, Golan, Dabbah, El-Yaniv, Puny, Galil, Moshe, Ronen, Nabwani et~al.}]{bercovich2025llama}
Akhiad Bercovich, Itay Levy, Izik Golan, Mohammad Dabbah, Ran El-Yaniv, Omri Puny, Ido Galil, Zach Moshe, Tomer Ronen, Najeeb Nabwani, and 1 others. 2025.
\newblock Llama-nemotron: Efficient reasoning models.
\newblock \emph{arXiv preprint arXiv:2505.00949}.

\bibitem[{Boyd and Kong(2017)}]{reasoning_boyd}
Maureen Boyd and Yiren Kong. 2017.
\newblock \href {https://doi.org/10.1080/0163853X.2015.1095596} {Reasoning words as linguistic features of exploratory talk: Classroom use and what it can tell us}.
\newblock \emph{Discourse Processes}, 54(1):62--81.

\bibitem[{Breidt(1996)}]{breidt1996extraction}
Elisabeth Breidt. 1996.
\newblock Extraction of vn-collocations from text corpora: A feasibility study for german.
\newblock \emph{arXiv preprint cmp-lg/9603006}.

\bibitem[{Bricken et~al.(2024)Bricken, Mishra-Sharma, Marcus, Jermyn, Olah, Rivoire, and Henighan}]{bricken2024stage}
Trenton Bricken, Siddharth Mishra-Sharma, Jonathan Marcus, Adam Jermyn, Christopher Olah, Kelley Rivoire, and Thomas Henighan. 2024.
\newblock Stage-wise model diffing.
\newblock \emph{Transformer Circuits Thread}.

\bibitem[{Bricken et~al.(2023)Bricken, Templeton, Batson, Chen, Jermyn, Conerly, Turner, Anil, Denison, Askell et~al.}]{bricken2023towards}
Trenton Bricken, Adly Templeton, Joshua Batson, Brian Chen, Adam Jermyn, Tom Conerly, Nick Turner, Cem Anil, Carson Denison, Amanda Askell, and 1 others. 2023.
\newblock Towards monosemanticity: Decomposing language models with dictionary learning, 2023.
\newblock \emph{URL https://transformer-circuits.pub/2023/monosemantic-features/index. html}, page~9.

\bibitem[{Brown et~al.(2020)Brown, Mann, Ryder, Subbiah, Kaplan, Dhariwal, Neelakantan, Shyam, Sastry, Askell et~al.}]{brown2020language}
Tom Brown, Benjamin Mann, Nick Ryder, Melanie Subbiah, Jared~D Kaplan, Prafulla Dhariwal, Arvind Neelakantan, Pranav Shyam, Girish Sastry, Amanda Askell, and 1 others. 2020.
\newblock Language models are few-shot learners.
\newblock \emph{Advances in neural information processing systems}, 33:1877--1901.

\bibitem[{Cao et~al.(2024)Cao, Zhang, Cao, Yin, Lin, Ma, and Chen}]{cao2024personalized}
Yuanpu Cao, Tianrong Zhang, Bochuan Cao, Ziyi Yin, Lu~Lin, Fenglong Ma, and Jinghui Chen. 2024.
\newblock Personalized steering of large language models: Versatile steering vectors through bi-directional preference optimization.
\newblock \emph{arXiv preprint arXiv:2406.00045}.

\bibitem[{Chanin et~al.(2024)Chanin, Wilken-Smith, Dulka, Bhatnagar, and Bloom}]{chanin2024absorption}
David Chanin, James Wilken-Smith, Tom{\'a}{\v{s}} Dulka, Hardik Bhatnagar, and Joseph Bloom. 2024.
\newblock A is for absorption: Studying feature splitting and absorption in sparse autoencoders.
\newblock \emph{arXiv preprint arXiv:2409.14507}.

\bibitem[{Chen et~al.(2021)Chen, Tworek, Jun, Yuan, Pinto, Kaplan, Edwards, Burda, Joseph, Brockman et~al.}]{chen2021evaluating}
Mark Chen, Jerry Tworek, Heewoo Jun, Qiming Yuan, Henrique Ponde De~Oliveira Pinto, Jared Kaplan, Harri Edwards, Yuri Burda, Nicholas Joseph, Greg Brockman, and 1 others. 2021.
\newblock Evaluating large language models trained on code.
\newblock \emph{arXiv preprint arXiv:2107.03374}.

\bibitem[{Chen et~al.(2023)Chen, Wu, Liang, Gong, Shou, Zhang, and Li}]{chen2023beyond}
Nuo Chen, Ning Wu, Shining Liang, Ming Gong, Linjun Shou, Dongmei Zhang, and Jia Li. 2023.
\newblock Beyond surface: Probing llama across scales and layers.
\newblock \emph{arXiv preprint arXiv:2312.04333}.

\bibitem[{Chinn and Anderson(1998)}]{structure_of_discussions}
Clark~A. Chinn and Richard~C. Anderson. 1998.
\newblock The structure of discussions that promote reasoning.
\newblock \emph{Teachers College Record}, 100(2):315--368.

\bibitem[{Conerly et~al.(2024)Conerly, Templeton, Bricken, Marcus, and Henighan}]{conerly2024update}
Tom Conerly, Adly Templeton, Trenton Bricken, Jonathan Marcus, and Tom Henighan. 2024.
\newblock Update on how we train saes.
\newblock \emph{Transformer Circuits Thread}.

\bibitem[{Cunningham et~al.(2023)Cunningham, Ewart, Riggs, Huben, and Sharkey}]{cunningham2023sae}
Hoagy Cunningham, Aidan Ewart, Logan Riggs, Robert Huben, and Lee Sharkey. 2023.
\newblock \href {https://arxiv.org/abs/2309.08600} {Sparse autoencoders find highly interpretable features in language models}.
\newblock \emph{Preprint}, arXiv:2309.08600.

\bibitem[{Cywi{\'n}ski and Deja(2025)}]{cywinski2025saeuron}
Bartosz Cywi{\'n}ski and Kamil Deja. 2025.
\newblock Saeuron: Interpretable concept unlearning in diffusion models with sparse autoencoders.
\newblock \emph{arXiv preprint arXiv:2501.18052}.

\bibitem[{Durmus et~al.(2024)Durmus, Tamkin, Clark, Wei, Marcus, Batson, Handa, Lovitt, Tong, McCain et~al.}]{durmusevaluating}
Esin Durmus, Alex Tamkin, Jack Clark, Jerry Wei, Jonathan Marcus, Joshua Batson, Kunal Handa, Liane Lovitt, Meg Tong, Miles McCain, and 1 others. 2024.
\newblock Evaluating feature steering: A case study in mitigating social biases, 2024.
\newblock \emph{URL https://anthropic. com/research/evaluating-feature-steering}.

\bibitem[{Elhage et~al.(2022)Elhage, Hume, Olsson, Schiefer, Henighan, Kravec, Hatfield-Dodds, Lasenby, Drain, Chen et~al.}]{elhage2022toy}
Nelson Elhage, Tristan Hume, Catherine Olsson, Nicholas Schiefer, Tom Henighan, Shauna Kravec, Zac Hatfield-Dodds, Robert Lasenby, Dawn Drain, Carol Chen, and 1 others. 2022.
\newblock Toy models of superposition.
\newblock \emph{arXiv preprint arXiv:2209.10652}.

\bibitem[{Gao et~al.(2024{\natexlab{a}})Gao, Dupr{\'e}~la Tour, Tillman, Goh, Troll, Radford, Sutskever, Leike, and Wu}]{gao2024scaling}
Leo Gao, Tom Dupr{\'e}~la Tour, Henk Tillman, Gabriel Goh, Rajan Troll, Alec Radford, Ilya Sutskever, Jan Leike, and Jeffrey Wu. 2024{\natexlab{a}}.
\newblock Scaling and evaluating sparse autoencoders.
\newblock \emph{arXiv preprint arXiv:2406.04093}.

\bibitem[{Gao et~al.(2024{\natexlab{b}})Gao, la~Tour, Tillman, Goh, Troll, Radford, Sutskever, Leike, and Wu}]{gao2024scalingSAE}
Leo Gao, Tom~Dupré la~Tour, Henk Tillman, Gabriel Goh, Rajan Troll, Alec Radford, Ilya Sutskever, Jan Leike, and Jeffrey Wu. 2024{\natexlab{b}}.
\newblock \href {https://arxiv.org/abs/2406.04093} {Scaling and evaluating sparse autoencoders}.
\newblock \emph{Preprint}, arXiv:2406.04093.

\bibitem[{Gerns and Mortimore(2025)}]{conginive_discource_explore}
Pilar Gerns and Louisa Mortimore. 2025.
\newblock Towards exploratory talk in secondary-school clil: An empirical study of the cognitive discourse function ‘explore’.
\newblock \emph{Language Teaching Research}.

\bibitem[{Grattafiori et~al.(2024)Grattafiori, Dubey, Jauhri, Pandey, Kadian, Al-Dahle, Letman, Mathur, Schelten, Vaughan et~al.}]{grattafiori2024llama}
Aaron Grattafiori, Abhimanyu Dubey, Abhinav Jauhri, Abhinav Pandey, Abhishek Kadian, Ahmad Al-Dahle, Aiesha Letman, Akhil Mathur, Alan Schelten, Alex Vaughan, and 1 others. 2024.
\newblock The llama 3 herd of models.
\newblock \emph{arXiv preprint arXiv:2407.21783}.

\bibitem[{Guo et~al.(2025)Guo, Yang, Zhang, Song, Zhang, Xu, Zhu, Ma, Wang, Bi et~al.}]{guo2025deepseek}
Daya Guo, Dejian Yang, Haowei Zhang, Junxiao Song, Ruoyu Zhang, Runxin Xu, Qihao Zhu, Shirong Ma, Peiyi Wang, Xiao Bi, and 1 others. 2025.
\newblock Deepseek-r1: Incentivizing reasoning capability in llms via reinforcement learning.
\newblock \emph{arXiv preprint arXiv:2501.12948}.

\bibitem[{Hendrycks et~al.(2021)Hendrycks, Burns, Kadavath, Arora, Basart, Tang, Song, and Steinhardt}]{hendrycks2021measuringmathematicalproblemsolving}
Dan Hendrycks, Collin Burns, Saurav Kadavath, Akul Arora, Steven Basart, Eric Tang, Dawn Song, and Jacob Steinhardt. 2021.
\newblock \href {https://arxiv.org/abs/2103.03874} {Measuring mathematical problem solving with the math dataset}.
\newblock \emph{Preprint}, arXiv:2103.03874.

\bibitem[{Huben et~al.(2024)Huben, Cunningham, Smith, Ewart, and Sharkey}]{huben2024sparse}
Robert Huben, Hoagy Cunningham, Logan~R. Smith, Aidan Ewart, and Lee Sharkey. 2024.
\newblock Sparse autoencoders find highly interpretable features in language models.
\newblock In \emph{Proceedings of the International Conference on Learning Representations (ICLR)}.

\bibitem[{Jiang et~al.(2024)Jiang, Rajendran, Ravikumar, Aragam, and Veitch}]{jiang2024origins}
Yibo Jiang, Goutham Rajendran, Pradeep Ravikumar, Bryon Aragam, and Victor Veitch. 2024.
\newblock On the origins of linear representations in large language models.
\newblock \emph{arXiv preprint arXiv:2403.03867}.

\bibitem[{Jin et~al.(2024)Jin, Yu, Huang, Zeng, Wang, Hua, Zhao, Mei, Meng, Ding et~al.}]{jin2024exploring}
Mingyu Jin, Qinkai Yu, Jingyuan Huang, Qingcheng Zeng, Zhenting Wang, Wenyue Hua, Haiyan Zhao, Kai Mei, Yanda Meng, Kaize Ding, and 1 others. 2024.
\newblock Exploring concept depth: How large language models acquire knowledge and concept at different layers?
\newblock \emph{arXiv preprint arXiv:2404.07066}.

\bibitem[{Kingma and Ba(2014)}]{kingma2014adam}
Diederik~P Kingma and Jimmy Ba. 2014.
\newblock Adam: A method for stochastic optimization.
\newblock \emph{arXiv preprint arXiv:1412.6980}.

\bibitem[{Kojima et~al.(2022)Kojima, Gu, Reid, Matsuo, and Iwasawa}]{kojima2022large}
Takeshi Kojima, Shixiang~Shane Gu, Machel Reid, Yutaka Matsuo, and Yusuke Iwasawa. 2022.
\newblock Large language models are zero-shot reasoners.
\newblock \emph{Advances in neural information processing systems}, 35:22199--22213.

\bibitem[{Kuznetsov et~al.(2025)Kuznetsov, Kushnareva, Druzhinina, Razzhigaev, Voznyuk, Piontkovskaya, Burnaev, and Barannikov}]{kuznetsov2025feature}
Kristian Kuznetsov, Laida Kushnareva, Polina Druzhinina, Anton Razzhigaev, Anastasia Voznyuk, Irina Piontkovskaya, Evgeny Burnaev, and Serguei Barannikov. 2025.
\newblock Feature-level insights into artificial text detection with sparse autoencoders.
\newblock \emph{arXiv preprint arXiv:2503.03601}.

\bibitem[{Leask et~al.(2025)Leask, Bussmann, Pearce, Bloom, Tigges, Al~Moubayed, Sharkey, and Nanda}]{leask2025canonical}
Patrick Leask, Bart Bussmann, Michael~T. Pearce, Joseph~I. Bloom, Curt Tigges, Noura Al~Moubayed, Lee Sharkey, and Neel Nanda. 2025.
\newblock Sparse autoencoders do not find canonical units of analysis.
\newblock In \emph{Proceedings of the International Conference on Learning Representations (ICLR)}.

\bibitem[{Lieberum et~al.(2024)Lieberum, Rajamanoharan, Conmy, Smith, Sonnerat, Varma, Kram{\'a}r, Dragan, Shah, and Nanda}]{lieberum2024gemma}
Tom Lieberum, Senthooran Rajamanoharan, Arthur Conmy, Lewis Smith, Nicolas Sonnerat, Vikrant Varma, J{\'a}nos Kram{\'a}r, Anca Dragan, Rohin Shah, and Neel Nanda. 2024.
\newblock Gemma scope: Open sparse autoencoders everywhere all at once on gemma 2.
\newblock \emph{arXiv preprint arXiv:2408.05147}.

\bibitem[{Lindsey et~al.(2024)Lindsey, Templeton, Marcus, Conerly, Batson, and Olah}]{lindsey2024sparse}
Jack Lindsey, Adly Templeton, Jonathan Marcus, Thomas Conerly, Joshua Batson, and Christopher Olah. 2024.
\newblock Sparse crosscoders for cross-layer features and model diffing.
\newblock \emph{Transformer Circuits Thread}.

\bibitem[{Madaan et~al.(2023)Madaan, Tandon, Gupta, Hallinan, Gao, Wiegreffe, Alon, Dziri, Prabhumoye, Yang et~al.}]{madaan2023self}
Aman Madaan, Niket Tandon, Prakhar Gupta, Skyler Hallinan, Luyu Gao, Sarah Wiegreffe, Uri Alon, Nouha Dziri, Shrimai Prabhumoye, Yiming Yang, and 1 others. 2023.
\newblock Self-refine: Iterative refinement with self-feedback.
\newblock \emph{Advances in Neural Information Processing Systems}, 36:46534--46594.

\bibitem[{Marks et~al.(2024)Marks, Rager, Michaud, Belinkov, Bau, and Mueller}]{marks2024sparse}
Samuel Marks, Can Rager, Eric~J Michaud, Yonatan Belinkov, David Bau, and Aaron Mueller. 2024.
\newblock Sparse feature circuits: Discovering and editing interpretable causal graphs in language models.
\newblock \emph{arXiv preprint arXiv:2403.19647}.

\bibitem[{Meng et~al.(2022)Meng, Bau, Andonian, and Belinkov}]{meng2022locating}
Kevin Meng, David Bau, Alex Andonian, and Yonatan Belinkov. 2022.
\newblock Locating and editing factual associations in gpt.
\newblock In \emph{Advances in Neural Information Processing Systems}.

\bibitem[{Michel et~al.(2011)Michel, Shen, Aiden, Veres, Gray, Team, Pickett, Hoiberg, Clancy, Norvig et~al.}]{michel2011quantitative}
Jean-Baptiste Michel, Yuan~Kui Shen, Aviva~Presser Aiden, Adrian Veres, Matthew~K Gray, Google~Books Team, Joseph~P Pickett, Dale Hoiberg, Dan Clancy, Peter Norvig, and 1 others. 2011.
\newblock Quantitative analysis of culture using millions of digitized books.
\newblock \emph{science}, 331(6014):176--182.

\bibitem[{Mihalcea and Tarau(2004)}]{mihalcea2004textrank}
Rada Mihalcea and Paul Tarau. 2004.
\newblock Textrank: Bringing order into text.
\newblock In \emph{Proceedings of the 2004 conference on empirical methods in natural language processing}, pages 404--411.

\bibitem[{Mikolov et~al.(2013)Mikolov, Chen, Corrado, and Dean}]{mikolov2013efficient}
Tomas Mikolov, Kai Chen, Greg Corrado, and Jeffrey Dean. 2013.
\newblock Efficient estimation of word representations in vector space.
\newblock \emph{arXiv preprint arXiv:1301.3781}.

\bibitem[{Muennighoff et~al.(2025)Muennighoff, Yang, Shi, Li, Fei-Fei, Hajishirzi, Zettlemoyer, Liang, Cand{\`e}s, and Hashimoto}]{muennighoff2025s1}
Niklas Muennighoff, Zitong Yang, Weijia Shi, Xiang~Lisa Li, Li~Fei-Fei, Hannaneh Hajishirzi, Luke Zettlemoyer, Percy Liang, Emmanuel Cand{\`e}s, and Tatsunori Hashimoto. 2025.
\newblock s1: Simple test-time scaling.
\newblock \emph{arXiv preprint arXiv:2501.19393}.

\bibitem[{Nanda et~al.(2023)Nanda, Lee, and Wattenberg}]{nanda2023emergent}
Neel Nanda, Andrew Lee, and Martin Wattenberg. 2023.
\newblock Emergent linear representations in world models of self-supervised sequence models.
\newblock In \emph{Proceedings of the 6th BlackboxNLP Workshop: Analyzing and Interpreting Neural Networks for NLP}, pages 16--30.

\bibitem[{Nostalgebraist(2020)}]{nostalgebraist2020logitlens}
Nostalgebraist. 2020.
\newblock \href {https://www.lesswrong.com/posts/AcKRB8wDpdaN6v6ru/interpreting-gpt-the-logit-lens} {Interpreting gpt: the logit lens}.

\bibitem[{of~America(2024)}]{aime}
Mathematical~Association of~America. 2024.
\newblock \href {https://artofproblemsolving.com/wiki/index.php/AIME_Problems_and_Solutions/} {Aime}.

\bibitem[{OpenAI(2024b)}]{openaio1}
OpenAI. 2024b.
\newblock Learning to reason with llms.
\newblock \url{https://openai.com/index/learning-to-reason-with-llms/}.

\bibitem[{OpenThoughts(2025)}]{openthoughts}
OpenThoughts. 2025.
\newblock {Open Thoughts}.
\newblock https://open-thoughts.ai.

\bibitem[{Park et~al.(2023)Park, Choe, and Veitch}]{park2023linear}
Kiho Park, Yo~Joong Choe, and Victor Veitch. 2023.
\newblock The linear representation hypothesis and the geometry of large language models.
\newblock \emph{arXiv preprint arXiv:2311.03658}.

\bibitem[{Paulo et~al.(2024)Paulo, Mallen, Juang, and Belrose}]{paulo2024automatically}
Gon{\c{c}}alo Paulo, Alex Mallen, Caden Juang, and Nora Belrose. 2024.
\newblock Automatically interpreting millions of features in large language models.
\newblock \emph{arXiv preprint arXiv:2410.13928}.

\bibitem[{Rayson and Garside(2000)}]{frequency_profiling}
Paul Rayson and Roger Garside. 2000.
\newblock \href {https://doi.org/10.3115/1117729.1117730} {Comparing corpora using frequency profiling}.
\newblock In \emph{Proceedings of the Workshop on Comparing Corpora - Volume 9}, WCC '00, pages 1--6, USA. Association for Computational Linguistics.

\bibitem[{Rein et~al.(2023)Rein, Hou, Stickland, Petty, Pang, Dirani, Michael, and Bowman}]{rein2023gpqagraduatelevelgoogleproofqa}
David Rein, Betty~Li Hou, Asa~Cooper Stickland, Jackson Petty, Richard~Yuanzhe Pang, Julien Dirani, Julian Michael, and Samuel~R. Bowman. 2023.
\newblock \href {https://arxiv.org/abs/2311.12022} {Gpqa: A graduate-level google-proof q\&a benchmark}.
\newblock \emph{Preprint}, arXiv:2311.12022.

\bibitem[{Shao et~al.(2024)Shao, Wang, Zhu, Xu, Song, Bi, Zhang, Zhang, Li, Wu et~al.}]{shao2024deepseekmath}
Zhihong Shao, Peiyi Wang, Qihao Zhu, Runxin Xu, Junxiao Song, Xiao Bi, Haowei Zhang, Mingchuan Zhang, YK~Li, Y~Wu, and 1 others. 2024.
\newblock Deepseekmath: Pushing the limits of mathematical reasoning in open language models.
\newblock \emph{arXiv preprint arXiv:2402.03300}.

\bibitem[{Shi et~al.(2025)Shi, Li, Liang, Wan, Ma, Wang, and He}]{shi2025routeSAE}
Wei Shi, Sihang Li, Tao Liang, Mingyang Wan, Guojun Ma, Xiang Wang, and Xiangnan He. 2025.
\newblock Route sparse autoencoder to interpret large language models.
\newblock \emph{arXiv preprint arXiv:2503.08200}.

\bibitem[{Shinn et~al.(2023)Shinn, Cassano, Gopinath, Narasimhan, and Yao}]{shinn2023reflexion}
Noah Shinn, Federico Cassano, Ashwin Gopinath, Karthik Narasimhan, and Shunyu Yao. 2023.
\newblock Reflexion: Language agents with verbal reinforcement learning.
\newblock \emph{Advances in Neural Information Processing Systems}, 36:8634--8652.

\bibitem[{Shu et~al.(2025)Shu, Wu, Zhao, Rai, Yao, Liu, and Du}]{shu2025survey}
Dong Shu, Xuansheng Wu, Haiyan Zhao, Daking Rai, Ziyu Yao, Ninghao Liu, and Mengnan Du. 2025.
\newblock A survey on sparse autoencoders: Interpreting the internal mechanisms of large language models.
\newblock \emph{arXiv preprint arXiv:2503.05613}.

\bibitem[{Simonyan et~al.(2014)Simonyan, Vedaldi, and Zisserman}]{SimonyanVZ13}
Karen Simonyan, Andrea Vedaldi, and Andrew Zisserman. 2014.
\newblock Deep inside convolutional networks: Visualising image classification models and saliency maps.
\newblock In \emph{ICLR (Workshop Poster)}.

\bibitem[{Soboleva et~al.(2023)Soboleva, Al-Khateeb, Myers, Steeves, Hestness, and Dey}]{cerebras2023slimpajama}
Daria Soboleva, Faisal Al-Khateeb, Robert Myers, Jacob~R Steeves, Joel Hestness, and Nolan Dey. 2023.
\newblock \href {https://huggingface.co/datasets/cerebras/SlimPajama-627B} {{SlimPajama: A 627B token cleaned and deduplicated version of RedPajama}}.
\newblock \url{https://www.cerebras.net/blog/slimpajama-a-627b-token-cleaned-and-deduplicated-version-of-redpajama}.

\bibitem[{Soo et~al.(2025)Soo, Teng, and Balaganesh}]{soo2025steering}
Samuel Soo, Wesley Teng, and Chandrasekaran Balaganesh. 2025.
\newblock Steering large language models with feature guided activation additions.
\newblock \emph{arXiv preprint arXiv:2501.09929}.

\bibitem[{Team(2024)}]{team2024qwq}
Qwen Team. 2024.
\newblock Qwq: Reflect deeply on the boundaries of the unknown.
\newblock \emph{Hugging Face}.

\bibitem[{Templeton(2024)}]{templeton2024scaling}
Adly Templeton. 2024.
\newblock \emph{Scaling monosemanticity: Extracting interpretable features from claude 3 sonnet}.
\newblock Anthropic.

\bibitem[{Vaswani et~al.(2017)Vaswani, Shazeer, Parmar, Uszkoreit, Jones, Gomez, Kaiser, and Polosukhin}]{vaswani2017attention}
Ashish Vaswani, Noam Shazeer, Niki Parmar, Jakob Uszkoreit, Llion Jones, Aidan~N Gomez, {\L}ukasz Kaiser, and Illia Polosukhin. 2017.
\newblock \href {http://arxiv.org/abs/1706.03762} {Attention is all you need}.
\newblock In \emph{Advances in neural information processing systems}, pages 5998--6008.

\bibitem[{Wang et~al.(2022)Wang, Wei, Schuurmans, Le, Chi, Narang, Chowdhery, and Zhou}]{wang2022self}
Xuezhi Wang, Jason Wei, Dale Schuurmans, Quoc Le, Ed~Chi, Sharan Narang, Aakanksha Chowdhery, and Denny Zhou. 2022.
\newblock Self-consistency improves chain of thought reasoning in language models.
\newblock \emph{arXiv preprint arXiv:2203.11171}.

\bibitem[{Wei et~al.(2022)Wei, Wang, Schuurmans, Bosma, Xia, Chi, Le, Zhou et~al.}]{wei2022chain}
Jason Wei, Xuezhi Wang, Dale Schuurmans, Maarten Bosma, Fei Xia, Ed~Chi, Quoc~V Le, Denny Zhou, and 1 others. 2022.
\newblock Chain-of-thought prompting elicits reasoning in large language models.
\newblock \emph{Advances in neural information processing systems}, 35:24824--24837.

\bibitem[{Yan et~al.(2024)Yan, Xiang, Chen, Wang, Gui, and He}]{yan2024monosemanticity}
Hanqi Yan, Yanzheng Xiang, Guangyi Chen, Yifei Wang, Lin Gui, and Yulan He. 2024.
\newblock Encourage or inhibit monosemanticity? revisiting monosemanticity from a feature decorrelation perspective.
\newblock \emph{arXiv:2406.17969}.

\bibitem[{Zhang et~al.(2020)Zhang, Li, Yue, and Yang}]{zhang2020empirical}
Mingxi Zhang, Xuemin Li, Shuibo Yue, and Liuqian Yang. 2020.
\newblock An empirical study of textrank for keyword extraction.
\newblock \emph{IEEE access}, 8:178849--178858.

\bibitem[{Zheng et~al.(2023)Zheng, Chiang, Sheng, Li, Zhuang, Wu, Zhuang, Li, Lin, Xing, Gonzalez, Stoica, and Zhang}]{zheng2023lmsyschat1m}
Lianmin Zheng, Wei-Lin Chiang, Ying Sheng, Tianle Li, Siyuan Zhuang, Zhanghao Wu, Yonghao Zhuang, Zhuohan Li, Zi~Lin, Eric.~P Xing, Joseph~E. Gonzalez, Ion Stoica, and Hao Zhang. 2023.
\newblock \href {https://arxiv.org/abs/2309.11998} {Lmsys-chat-1m: A large-scale real-world llm conversation dataset}.
\newblock \emph{Preprint}, arXiv:2309.11998.

\end{thebibliography}
